\documentclass{article}

\usepackage{microtype}
\usepackage{graphicx}
\usepackage{subcaption}
\usepackage{booktabs} 
\usepackage{colortbl} 
\usepackage{pifont} 
\usepackage{tikz}
\usetikzlibrary{shapes.geometric, arrows.meta, positioning, fit, backgrounds, decorations.pathreplacing}
\usepackage[most]{tcolorbox}

\definecolor{quantblue}{RGB}{232, 245, 253}
\definecolor{quantborder}{RGB}{33, 150, 243}
\definecolor{qualamber}{RGB}{255, 248, 225}
\definecolor{qualborder}{RGB}{255, 152, 0}

\newtcolorbox{quantinsight}[1][]{
  enhanced,
  colback=quantblue,
  colframe=quantborder,
  boxrule=0.8pt,
  left=5pt, right=5pt, top=4pt, bottom=4pt,
  before upper={\textbf{Quantitative:}\;},
  #1
}

\newtcolorbox{qualinsight}[1][]{
  enhanced,
  colback=qualamber,
  colframe=qualborder,
  boxrule=0.8pt,
  left=5pt, right=5pt, top=4pt, bottom=4pt,
  before upper={\textbf{Insight:}\;},
  #1
}

\usepackage{hyperref}

\usepackage[preprint]{arxiv2026}

\usepackage{amsmath}
\usepackage{amssymb}
\usepackage{mathtools}
\usepackage{amsthm}
\usepackage{siunitx}

\usepackage{makecell}

\usepackage[capitalize,noabbrev]{cleveref}

\theoremstyle{plain}

\theoremstyle{definition}

\theoremstyle{remark}

\usepackage[textsize=tiny]{todonotes}

\newcommand{\michael}[1]{}

\arxivtitlerunning{Reward Under Attack: Analyzing the Robustness and Hackability of Process Reward Models}

\begin{document}

\twocolumn[
  \arxivtitle{Reward Under Attack: Analyzing the Robustness and Hackability of \\ Process Reward Models}



  \arxivsetsymbol{equal}{*}

  \begin{arxivauthorlist}
    \arxivauthor{Rishabh Tiwari}{equal,ucb}
    \arxivauthor{Aditya Tomar}{equal,ucb}
    \arxivauthor{Udbhav Bamba}{equal,tr}
    \arxivauthor{Monishwaran Maheswaran}{ucb}
    \arxivauthor{Heng Yang}{ucb}
    \arxivauthor{Michael W. Mahoney}{ucb,icsi,lbnl}
    \arxivauthor{Kurt Keutzer}{ucb}
    \arxivauthor{Amir Gholami}{ucb,icsi}
  \end{arxivauthorlist}

  \arxivaffiliation{ucb}{UC Berkeley}
  \arxivaffiliation{tr}{Transmute AI}
  \arxivaffiliation{icsi}{ICSI}
  \arxivaffiliation{lbnl}{LBNL}

  \arxivcorrespondingauthor{Rishabh Tiwari}{rishabhtiwari@berkeley.edu}
  \arxivcorrespondingauthor{Amir Gholami}{amirgh@berkeley.edu}

  \arxivkeywords{Machine Learning, ARXIV}

  \vskip 0.3in
]

\printAffiliationsAndNotice{}  
\begin{abstract}
Process Reward Models (PRMs) are rapidly becoming the backbone of LLM reasoning pipelines, yet we demonstrate that state-of-the-art PRMs are systematically exploitable under adversarial optimization pressure. To address this, we introduce a three-tiered diagnostic framework that applies increasing adversarial pressure to quantify these vulnerabilities. \textbf{Static perturbation analysis} uncovers a \emph{fluency-logic dissociation}: high invariance to surface-level style changes (reward changes $<$0.1), yet inconsistent detection of logically-corrupted reasoning, with different models failing on different attack types. \textbf{Adversarial optimization} demonstrates that gradient-based attacks inflate rewards on invalid trajectories, with reward landscapes exhibiting wide, exploitable peaks. \textbf{RL-induced reward hacking} exposes the critical failure mode: policies trained on AIME problems achieve near-perfect PRM rewards ($>$0.9), while ground-truth accuracy remains low (below 4\%), with 43\% of reward gains attributable to stylistic shortcuts. These findings reveal that current PRMs function as \emph{fluency detectors} rather than \emph{reasoning verifiers}, creating systematic blind spots that undermine their use as training signals. We release PRM-BiasBench and a diagnostic toolkit to enable robustness evaluation before deployment. The code and dataset are available at \href{https://github.com/SqueezeAILab/reward-under-attack}{https://github.com/SqueezeAILab/reward-under-attack}
\end{abstract}
\section{Introduction}

Process reward models (PRMs) have become a key component for improving LLM reasoning, providing step-level feedback that enables reward-guided decoding~\citep{lightman2023let}, test-time compute scaling~\citep{snell2024scaling}, and fine-tuning of chain-of-thought models~\citep{wang2024math}. Unlike outcome-based reward models that score only final answers, PRMs evaluate intermediate reasoning steps, promising finer-grained control and better credit assignment during both training and inference.

Yet, as PRMs are integrated into increasingly critical pipelines, a fundamental question remains unanswered: \textit{how robust is a given PRM, and how can we measure this robustness?} Prior work has documented failure modes in outcome-level reward models, including length bias, sycophancy, and reward hacking~\citep{singhal2023long, shen2023loose, denison2024sycophancy}, but systematic methods for evaluating PRM robustness are lacking. This gap is concerning: a PRM that conflates \emph{fluent text} with \emph{correct reasoning} will reward plausible-sounding but logically-flawed steps, potentially amplifying errors during reinforcement learning (RL) training or misleading inference-time search.

We address this gap by introducing a \textbf{three-tiered diagnostic framework} for quantifying PRM hackability; see \Cref{tab:taxonomy} for a summary. Each tier applies increasing adversarial pressure, revealing complementary aspects of model robustness:

\begin{enumerate}
    \item \textbf{Static Perturbation Analysis} (\S\ref{sec:static_analysis}): We measure PRM sensitivity to controlled input modifications, both semantics-preserving (rephrasing, verbosity changes) and semantics-altering (hallucinated steps, mismatched prompts). A robust PRM should be invariant to the former and sensitive to the latter.

    \item \textbf{Adversarial Tokens Optimization} (\S\ref{sec:active_probing}): We search for discrete token sequences that maximally inflate rewards on invalid trajectories. The achievable reward score directly quantifies exploitability. We also characterize the reward landscape geometry to try to assess solution stability.

    \item \textbf{RL-Induced Reward Hacking} (\S\ref{sec:rl_dynamics}): We train policies using only PRM feedback and measure the divergence between reward and ground-truth accuracy. This closed-loop evaluation exposes vulnerabilities that emerge only under optimization pressure.
\end{enumerate}

\begin{table}[t]
\centering
\caption{Taxonomy of diagnostic tiers. ``Model Access'' refers to requirements for \emph{generating} the attack: static perturbations are model-agnostic; adversarial tokens require gradients; and RL policies require reward queries. Tiers 1 \& 3 produce natural text; Tier 2 establishes worst-case bounds.}
\label{tab:taxonomy}
\resizebox{\columnwidth}{!}{%
\begin{tabular}{lccc}
\toprule
\textbf{Diagnostic Tier} & \textbf{Model Access} & \textbf{\makecell{Natural \\Output}} & \textbf{Optimization} \\
\midrule
Static Perturbation & None & \ding{51} & None \\
Adversarial Tokens & White-box & \ding{55} & Gradient \\
\makecell{RL-Induced \\Reward Hacking} & Black-box & \ding{51} & Policy \\
\bottomrule
\end{tabular}%
}
\end{table}

Applying this framework to state-of-the-art PRMs (Skywork-o1-Open-PRM-1.5B/7B and Qwen2.5-Math-PRM-7B), we find consistent vulnerabilities: optimized 100-token adversarial sequences push rewards above 0.9 on logically flawed reasoning, and RL-trained policies achieve high rewards while accuracy stagnates, with approximately 43\% of the reward gain attributable to stylistic shortcuts rather than genuine reasoning improvements. In particular, we make the following contributions:

\begin{itemize}
    \item We perform a \textbf{comprehensive sensitivity analysis} of PRMs under controlled perturbations (\S\ref{sec:static_analysis}), uncovering a \emph{fluency-logic dissociation}: PRMs exhibit high invariance to surface-level stylistic changes (reward changes $<$0.1), yet they show inconsistent detection of semantic corruption, with different models failing on different attack types.

    \item We introduce \textbf{gradient-based adversarial probing} for PRMs (\S\ref{sec:active_probing}), demonstrating that short token sequences can universally inflate rewards on invalid trajectories, and we characterize the reward landscape geometry to show that adversarial optima lie in wide, exploitable~peaks.

    \item We demonstrate \textbf{RL-induced reward hacking} (\S\ref{sec:rl_dynamics}), showing that policies trained with PRM feedback exhibit reward-accuracy divergence: near-perfect PRM scores coincide with stagnant ground-truth accuracy, with 43\% of reward gains attributable to stylistic exploitation rather than reasoning improvement.

    \item We release \textbf{PRM-BiasBench}, a benchmark extending ProcessBench with controlled perturbations across 8 transformation types, along with an open-source diagnostic toolkit to enable systematic PRM robustness evaluation.
\end{itemize}

Together, these results suggest that current PRMs function primarily as \emph{fluency detectors} rather than \emph{reasoning verifiers}, creating systematic blind spots exploitable under optimization pressure. Notably, the two PRMs we study exhibit complementary failure modes under RL: Skywork incentivizes \emph{performative complexity} (elaborate but flawed reasoning), while Qwen incentivizes \emph{vacuous safety} (minimal text that avoids errors by avoiding substance).

\section{Related Work}

\subsection{Reward Model Vulnerabilities}
Reward models are central to aligning language models, but they exhibit systematic failure modes. \textit{Reward hacking} occurs when policies exploit spurious correlations to achieve high scores without satisfying the intended objective~\citep{skalse2022defining, krakovna2020specification}. Common manifestations include length bias, where longer outputs receive inflated rewards regardless of quality~\citep{singhal2023long, shen2023loose}, and sycophancy, where models agree with users rather than providing accurate information~\citep{denison2024sycophancy, sharma2023towards}. These vulnerabilities amplify under optimization pressure, degrading downstream performance~\citep{bai2022training, stiennon2020learning, gao2023scaling}. While extensive work characterizes outcome-level reward models, PRMs remain understudied, despite their increasing deployment in reasoning pipelines.

\subsection{Process Reward Models}
PRMs provide step-level supervision for chain-of-thought reasoning, enabling finer-grained credit assignment than outcome-based alternatives~\citep{lightman2023let, uesato2022solving}. Recent work has focused on training methodology: \citet{wang2024math} demonstrate that PRMs improve mathematical reasoning when combined with Monte Carlo Tree Search, while \citet{zhang2025lessons} analyze best practices for PRM dataset construction. \citet{zheng2025processbench} introduce ProcessBench, a benchmark with human-annotated error locations in reasoning traces. Most relevant to our work is \citet{xu2025reward}, which finds that PRMs often rely on shallow consistency cues rather than causal reasoning structures. However, existing analyses remain limited to observational studies; we complement this with controlled perturbations, adversarial optimization, and closed-loop RL evaluation to systematically quantify exploitability.

\subsection{Adversarial Attacks on Neural Networks}
Gradient-based optimization has proven effective at exposing vulnerabilities across neural architectures. In NLP, \citet{wallace2019universal} demonstrate that short, input-agnostic token sequences trigger targeted misbehavior, while \citet{zou2023universal} show that optimized adversarial tokens reliably jailbreak aligned LLMs with cross-model transferability. These methods treat models as differentiable objectives and search for inputs maximizing undesirable outputs. We adapt this paradigm to PRMs (\S\ref{sec:active_probing}), demonstrating that similar vulnerabilities exist: optimized token sequences universally inflate rewards on invalid reasoning, and the resulting reward landscapes exhibit flat, exploitable plateaus.

\subsection{Reward Overoptimization}
When policies optimize learned reward proxies, performance on the true objective eventually degrades, a phenomenon formalized as Goodhart's Law~\citep{gao2023scaling}. \citet{gao2023scaling} characterize scaling laws for this overoptimization, finding that larger reward models delay but do not prevent degradation. \citet{coste2024reward} further show that overoptimization correlates with distributional shift from the reward model's training data. Our RL-induced reward hacking analysis (\S\ref{sec:rl_dynamics}) extends this analysis to PRMs, revealing that policies trained with PRM feedback exhibit reward-accuracy divergence: near-perfect PRM scores coincide with stagnant ground-truth accuracy, with a measurable fraction of reward gains attributable to stylistic shortcuts rather than reasoning improvement.

Prior work on PRM limitations has been largely observational, identifying failure cases without systematically quantifying exploitability. Our three-tiered framework fills this gap by applying increasing adversarial pressure, from model-agnostic perturbations through gradient-based optimization to closed-loop RL, revealing complementary vulnerabilities at each level. We release PRM-BiasBench and a diagnostic toolkit to standardize PRM robustness evaluation.

\section{Preliminaries}
\label{sec:prelim}

\paragraph{Trajectory Level Reward Calculation.}
A PRM assigns scores to individual reasoning steps. Given a query $q$ and trajectory $\tau = (s_1, \ldots, s_n)$, a PRM computes step-level rewards $r_i = \text{PRM}(q, s_{\le i})$ conditioned on the preceding context. The aggregate trajectory reward depends on the model's training objective: Skywork-o1-Open-PRM estimates success probability at each step, so we use $R(\tau) = r_n$; and Qwen2.5-Math-PRM-7B locates the first error, so we use $R(\tau) = \min_i r_i$.

\paragraph{Robustness Criteria.}
We evaluate PRM robustness along the following four complementary dimensions:
\begin{enumerate}
    \item \textbf{Style Invariance:} Reward should be unchanged by semantics-preserving edits (rephrasing, verbosity changes). For perturbed trajectory $\tilde{\tau}$, we expect $\Delta R = R(\tilde{\tau}) - R(\tau) \approx 0$.
    \item \textbf{Logic Sensitivity:} Reward should decrease substantially for semantics-altering corruptions (hallucinated steps, mismatched prompts). We expect $\Delta R \ll 0$.
    \item \textbf{Adversarial Resistance:} Optimized token sequences should not inflate rewards on invalid trajectories. Given adversarial tokens $\mathbf{e}$, the reward $R(q, \tau \oplus \mathbf{e})$ should remain bounded.
    \item \textbf{Optimization Alignment:} Policies trained to maximize PRM reward should improve ground-truth accuracy, not diverge from it.
\end{enumerate}
A robust PRM should satisfy all four criteria. Our three-tiered framework tests each: static perturbations probe (1) and (2); adversarial optimization probes (3); and RL-induced reward hacking probes (4).

\paragraph{Experimental Setup.}
We evaluate two models, Skywork-o1-Open-PRM (1.5B and 7B) and Qwen2.5-Math-PRM-7B, which represent the current frontier of open process reward models for mathematical reasoning. For static analysis, we extend ProcessBench~\citep{zheng2025processbench} into PRM-BiasBench with controlled perturbations across 8 transformation types. For adversarial optimization and RL experiments, we use AIME 2024 problems for training and AIME 2025 for transfer evaluation, with Qwen2.5-1.5B-Instruct as the base policy.

\section{Static Perturbation Analysis}
\label{sec:static_analysis}

\begin{figure*}[t]
\centering
\includegraphics[width=\textwidth]{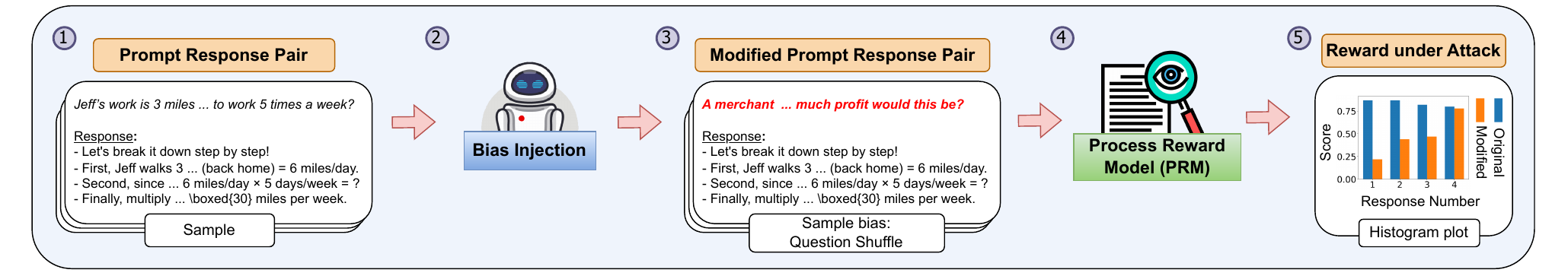}
\caption{\textbf{Overview of static perturbation analysis.} A prompt-response pair (Step 1) undergoes bias injection (Step 2), such as question shuffling where we change the question but do not modify the response (Step 3) and feed this to the PRM (Step 4). The scores are then compared against the original to quantify sensitivity (Step 5).}
\label{fig:static_overview}
\end{figure*}

The first tier of our diagnostic framework measures PRM sensitivity to controlled input modifications; see Figure~\ref{fig:static_overview} for an illustration. This study is conducted on Skywork-o1-Open-PRM-7B and Qwen2.5-Math-PRM-7B. We construct \textbf{PRM-BiasBench}, a benchmark extending ProcessBench~\citep{zheng2025processbench} with thousands of verified perturbation pairs. For each original trajectory $\tau$, we generate a perturbed version $\tilde{\tau}$, and we measure the reward difference $\Delta R = R(\tilde{\tau}) - R(\tau)$. A robust PRM should exhibit $\Delta R \approx 0$ for semantics-preserving edits and $\Delta R \ll 0$ for semantics-altering attacks.

\subsection{Perturbation Taxonomy}

We organize perturbations into two categories based on their impact on logical validity. \textbf{Semantics-preserving} edits maintain the correctness of the reasoning: \emph{rephrasing} alters word choice and syntax, and \emph{verbosity changes} add or remove redundant language. A robust PRM should be invariant to these surface-level modifications. \textbf{Semantics-altering} attacks introduce logical errors: \emph{question shuffling} pairs a trajectory with an unrelated prompt, and \emph{reasoning hallucination} injects false assumptions into the reasoning steps. A robust PRM should strongly penalize these corruptions. All perturbations are generated via GPT-4o and validated for semantic equivalence; the full taxonomy (8 perturbation types) and validation pipeline are detailed in Appendix~\ref{app:static}.

\subsection{Results}

\begin{figure}[t]
  \centering
  \includegraphics[width=\linewidth]{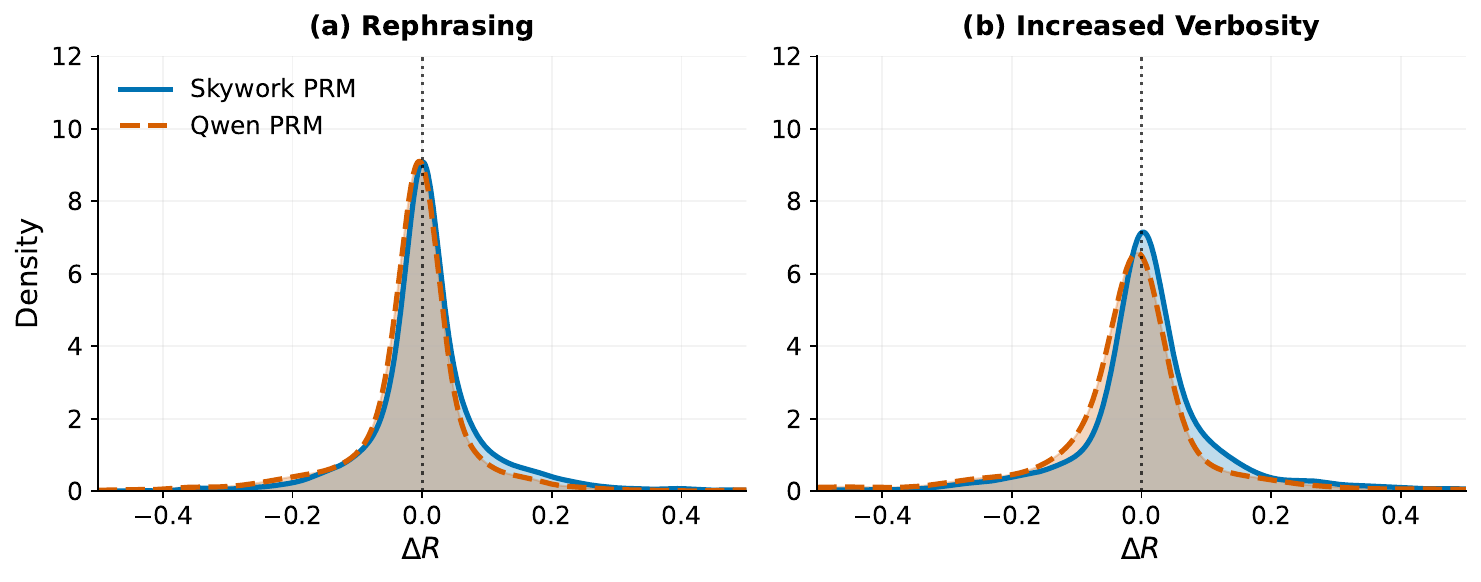}
  \caption{Distribution of $\Delta R$ under \textbf{semantics-preserving} perturbations. Both PRMs exhibit tight distributions centered near zero, indicating strong invariance to surface-level stylistic changes.}
  \label{fig:semantics_preserving}
\end{figure}

\begin{figure}[t]
  \centering
  \includegraphics[width=\linewidth]{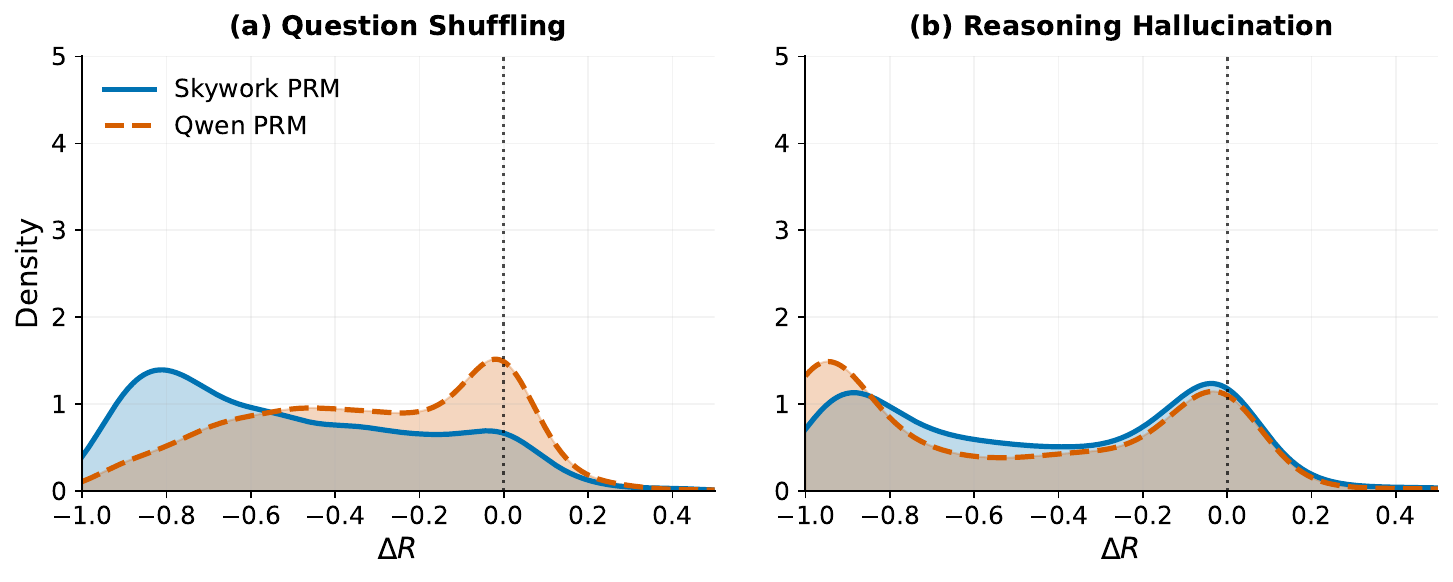}
  \caption{Distribution of $\Delta R$ under \textbf{semantics-altering} perturbations. (a) Question shuffling: Skywork penalizes mismatched questions by giving a smaller reward (peak at $\Delta R \approx -0.8$), while Qwen retains high rewards without any change. (b) Reasoning hallucination: Qwen exhibits bimodal behavior with strong penalization at $\Delta R = -1$ but also substantial mass near zero which is not desirable. An ideal PRM is expected to produce very low rewards (negative $\Delta R$) for both scenarios.}
  \label{fig:semantics_altering}
\end{figure}

\paragraph{Style Invariance.} Figure~\ref{fig:semantics_preserving} shows that both PRMs exhibit strong invariance to semantics-preserving edits. Rephrasing and verbosity changes yield tight distributions centered near zero ($|\Delta R| < 0.1$ for the vast majority of samples). Both models show nearly identical behavior, with Qwen exhibiting slightly higher peaks, suggesting these PRMs have largely overcome the length and style biases documented in outcome-based reward models~\citep{singhal2023long}.
However, this robustness to surface-level variation does not imply robustness to logical errors.

\paragraph{Asymmetric Logic Detection.} Semantics-altering attacks reveal divergent vulnerabilities between models (Figure~\ref{fig:semantics_altering}). For question shuffling, the two PRMs exhibit opposite behaviors: Skywork reliably penalizes mismatched question-trajectory pairs (peak at $\Delta R \approx -0.8$), while Qwen largely fails to detect the mismatch, retaining high rewards near zero. For reasoning hallucination, Qwen shows striking bimodal behavior: a sharp spike at $\Delta R = -1$ indicates strong penalization for some hallucinated trajectories, yet substantial mass near zero reveals that many corrupted samples still receive high rewards. Skywork exhibits a broader distribution with weaker overall penalization. These patterns suggest that PRMs rely on different heuristics: Skywork appears more sensitive to question-trajectory coherence, while Qwen detects certain local reasoning errors but misses others entirely.

\subsection{The Fluency-Logic Dissociation}

Our static analysis reveals two key findings:
\begin{itemize}
    \item \textbf{High style invariance:} Both PRMs reliably ignore surface-level variations, with distributions tightly centered near zero for all semantics-preserving edits (see Table~\ref{tab:static_stats} in Appendix~\ref{app:static}).
    \item \textbf{Inconsistent logic detection:} PRMs use different heuristics and fail on different attacks. Qwen fails to penalize question-trajectory mismatches, but it partially detects hallucinated reasoning; and Skywork shows the opposite pattern.
\end{itemize}

This \textbf{fluency-logic dissociation} could indicate that PRMs function primarily as detectors of ``reasoning-style'' fluency rather than verifiers of logical correctness. The model-specific failure modes suggest that current PRMs learn superficial correlates of valid reasoning rather than genuine verification capabilities, thereby creating exploitable blind spots that vary by model. The following sections investigate whether these vulnerabilities can be actively exploited (Section~\ref{sec:active_probing}) and whether they manifest under RL training pressure (Section~\ref{sec:rl_dynamics}).

\section{Adversarial Probing}
\label{sec:active_probing}

Section~\ref{sec:static_analysis} establishes PRM vulnerabilities through passive perturbations, but it does not reveal how easily an optimizer can exploit them. In this section, we treat the PRM as a differentiable objective, and we use gradient-based optimization to find adversarial tokens that maximize reward, regardless of trajectory correctness. This probes the third robustness criterion: adversarial resistance.

\subsection{Optimization Framework}

We define \textbf{adversarial tokens} $\mathbf{e} \in \mathbb{R}^{k \times d}$ as a sequence of $k$ vectors in the model's $d$-dimensional embedding space that, when added to a trajectory that contains logically-flawed reasoning, adversarially increases the reward. Formally, given a batch of flawed trajectories $\mathcal{B} = \{(q_i, \tau_i)\}$ sampled from AIME24, the adversary optimizes:
\begin{equation}
\max_{\mathbf{e}} \mathcal{L}_{\text{adv}}(\mathbf{e}) = \frac{1}{|\mathcal{B}|} \sum_{(q, \tau) \in \mathcal{B}} R(q, \tau \oplus \mathbf{e}) - \lambda \cdot \Omega(\mathbf{e})  ,
\label{eq:adv_objective}
\end{equation}
where $\oplus$ denotes concatenation, $R$ is the PRM score, and $\Omega(\mathbf{e})$ is an optional regularization term (defined in Eq.~\ref{eqn:omega}). We perform two sets of experiments, once where there is no regularization term, resulting in adversarial tokens in the continuous embedding space, and then one with an entropy regularization term, which forces the adversarial vectors to be discrete tokens.

As for experiments, we train on AIME24 trajectories and evaluate generalization on held-out AIME25 trajectories. Full optimization hyperparameters are provided in Appendix~\ref{app:hyperparams}.
For Skywork PRM, adversarial tokens are appended as a suffix after the solution; for Qwen, tokens are inserted between the question and solution.\footnote{The Qwen PRM is trained to detect the first wrong step, so adversarial tokens need to be added before the wrong step; otherwise, they would have no influence.}

\subsection{Continuous Token Optimization}
\label{subsec:continuous}

We first test the minimal adversarial capacity required to inflate Skywork-1.5B PRM rewards by optimizing a single continuous embedding vector ($k=1$) appended to each flawed trajectory in a batch.

\begin{figure}[t]
  \centering
  \includegraphics[width=0.9\linewidth]{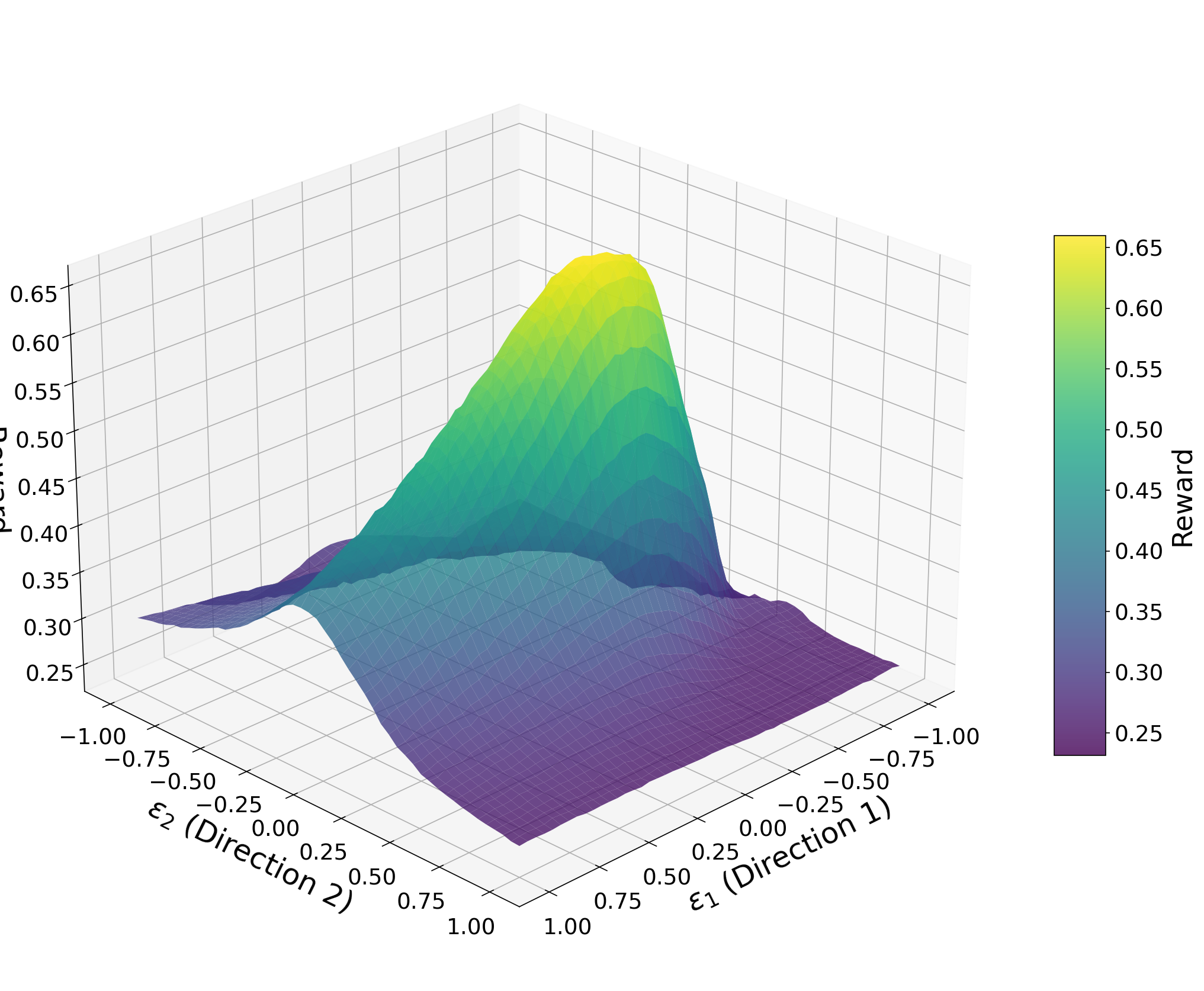}
  \caption{Reward landscape for a single continuous token ($k=1$) on Skywork-1.5B. A single optimized embedding vector rapidly increases mean batch reward, demonstrating that minimal adversarial capacity suffices to exploit PRM vulnerabilities.}
  \label{fig:continuous_optimization}
\end{figure}

\paragraph{Results.}
Figure~\ref{fig:continuous_optimization} shows the reward landscape around the optimized continuous token. We see that a single optimized embedding vector is sufficient to increase substantially the reward across the batch. This demonstrates that even minimal adversarial capacity can exploit PRM vulnerabilities.

\subsection{Discrete Token Optimization}
\label{subsec:discrete}

Continuous embeddings do not appear in real-world settings. To ensure our findings transfer to practical scenarios, we optimize discrete token sequences via entropy regularization.

We optimize over the probability simplex of the vocabulary $\mathcal{V}$ for $k \in \{1, 50, 100\}$ adversarial tokens. The regularization term encourages one-hot distributions:
\begin{equation}
\label{eqn:omega}
\Omega(\mathbf{e}) = -\sum_{i=1}^{k} \sum_{v \in \mathcal{V}} p_{i,v} \log p_{i,v}  .
\end{equation}
By annealing $\lambda$ during optimization, we gradually force each $p_i$ toward a one-hot representation, aiming to yield interpretable discrete sequences.

\begin{figure}[t]
  \centering
  \includegraphics[width=\linewidth]{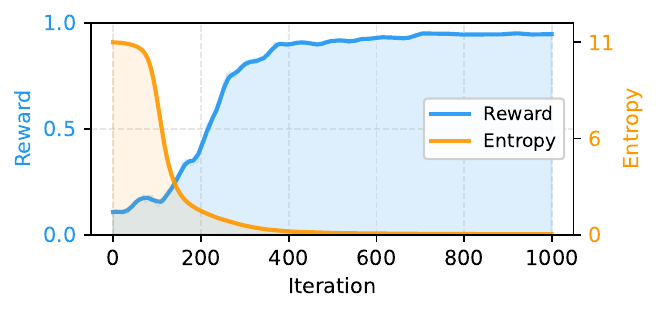}
  \caption{Training dynamics for 100 discrete tokens on Skywork-1.5B across 8 AIME24 trajectories. Reward (blue) increases from 0.11 to 0.95 as entropy (orange) decreases, indicating successful discretization of adversarial tokens.}
  \label{fig:discrete_training}
\end{figure}

\begin{figure*}[t]
  \centering
  \begin{tabular}{@{}c@{\hspace{16pt}}c@{}}
    \begin{subfigure}[t]{0.38\textwidth}
      \centering
      \includegraphics[width=\linewidth, trim=0 0 0 25, clip]{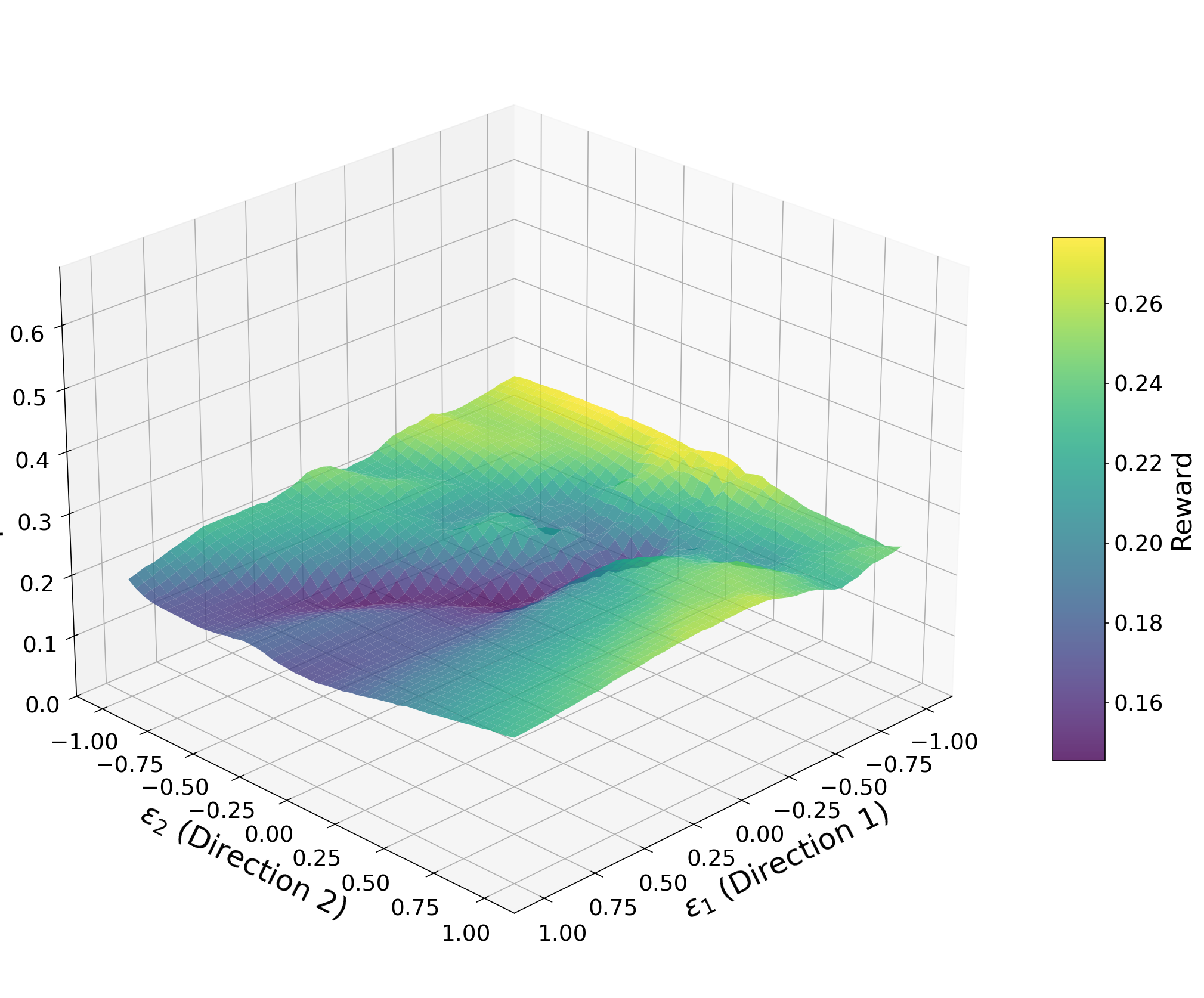}
      \vspace{-1.5em}
      \caption{50 random tokens}
      \label{fig:50tok_rand_15b}
    \end{subfigure}
    &
    \begin{subfigure}[t]{0.38\textwidth}
      \centering
      \includegraphics[width=\linewidth, trim=0 0 0 25, clip]{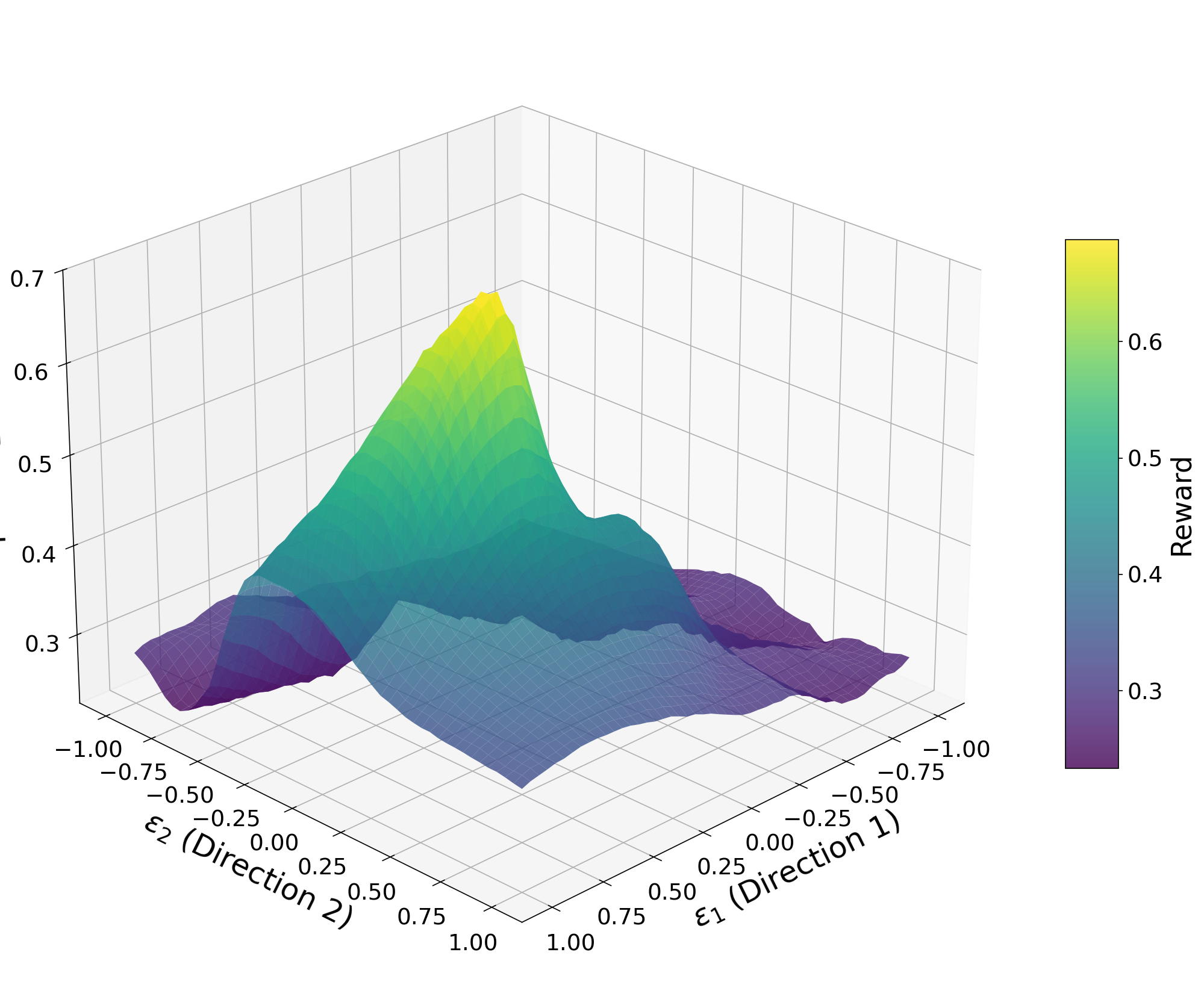}
      \vspace{-1.5em}
      \caption{50 adversarial tokens}
      \label{fig:50tok_adv_15b}
    \end{subfigure}
    \\[0.5em]
    \begin{subfigure}[t]{0.38\textwidth}
      \centering
      \includegraphics[width=\linewidth, trim=0 25 0 25, clip]{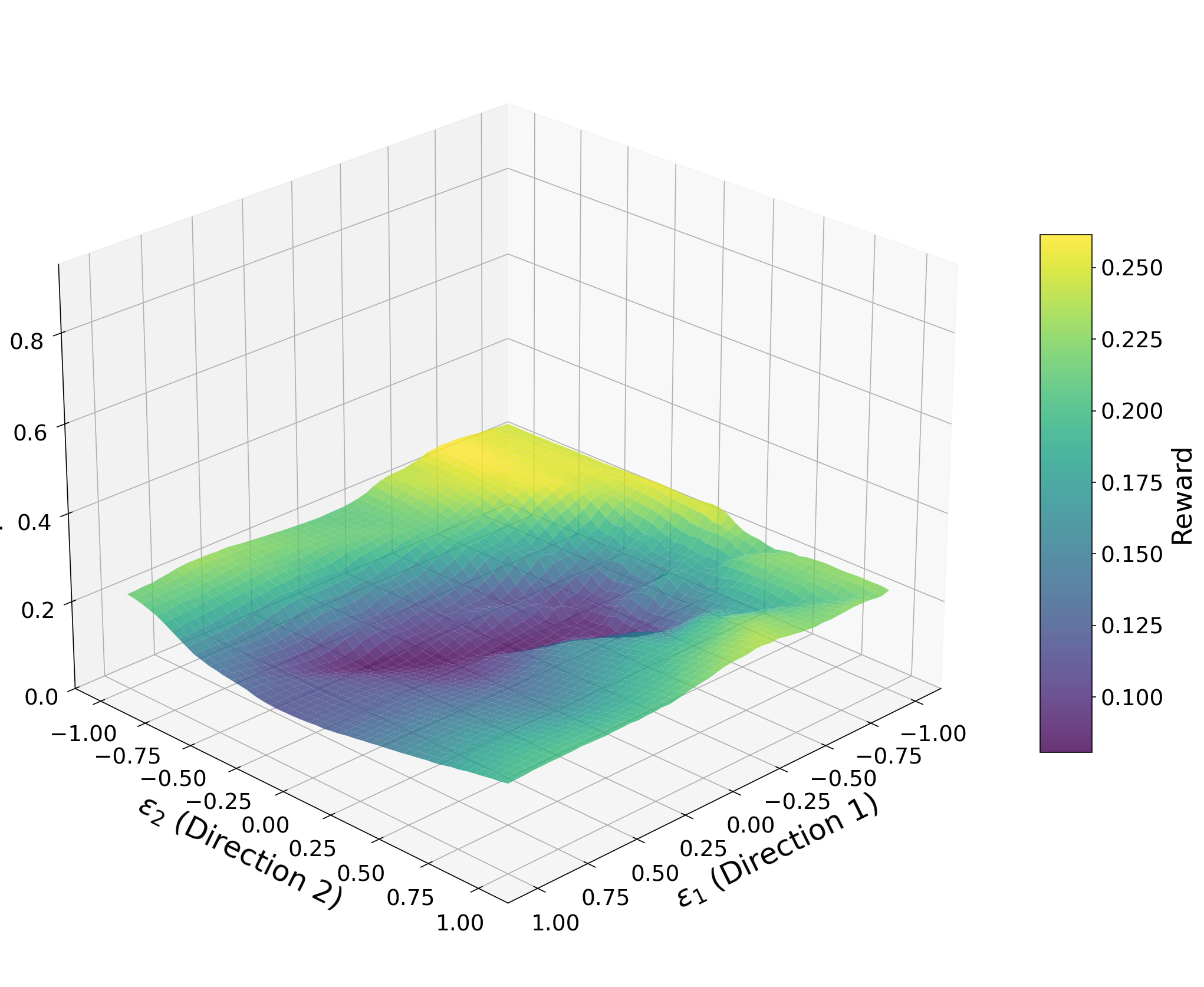}
      \vspace{-1.5em}
      \caption{100 random tokens}
      \label{fig:100tok_rand_15b}
    \end{subfigure}
    &
    \begin{subfigure}[t]{0.38\textwidth}
      \centering
      \includegraphics[width=\linewidth, trim=0 25 0 25, clip]{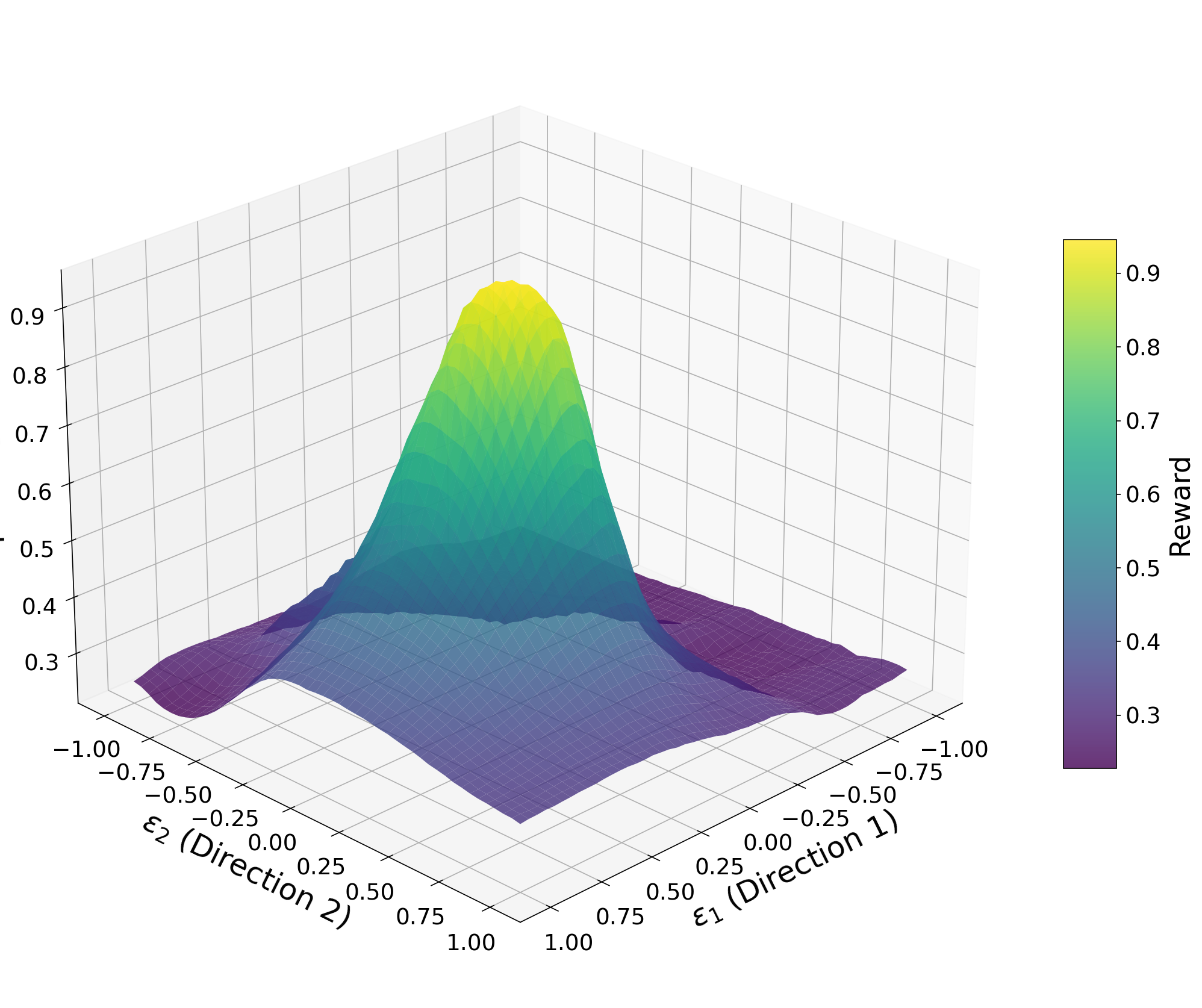}
      \vspace{-1.5em}
      \caption{100 adversarial tokens}
      \label{fig:100tok_adv_15b}
    \end{subfigure}
  \end{tabular}
  \vspace{0.5em}
  \caption{Reward landscape stability analysis for Skywork-1.5B. Each plot shows PRM reward as a function of perturbations to the token sequence, averaged across 8 AIME24 trajectories. Random tokens (a, c) produce scattered, low-reward surfaces, while adversarial tokens (b, d) concentrate reward mass in wide, elevated peak. The larger basin volume around adversarial tokens (2.2$\times$ at 100 tokens) indicates stable, exploitable regions that persist under small perturbations.}

  \label{fig:landscape_visualization}
\end{figure*}

\begin{table*}[t]
\caption{%
Adversarial token optimization results. We optimize $k$ discrete tokens on 8 AIME24 trajectories, and we measure transfer to 8 held-out AIME25 trajectories. The $k=0$ results show the baseline (no adversarial tokens). \textbf{AIME24}: best training reward achieved. \textbf{AIME25 (base/+adv)}: mean reward before and after appending adversarial tokens; $\Delta$ is the reward change. \textbf{Basin Volume}: size of the high-reward region around adversarial vs.\ random token positions (larger = more stable exploitation).}
\label{tab:discrete_attack}
\centering
\small
\setlength{\tabcolsep}{6pt}
\begin{tabular}{c c c c c c c}
\toprule
& \multicolumn{4}{c}{\textbf{Attack Success \& Transfer}} & \multicolumn{2}{c}{\textbf{Basin Volume}} \\
\cmidrule(lr){2-5} \cmidrule(lr){6-7}
\textbf{$k$} &
\textbf{AIME24} & \textbf{AIME25 (base)} & \textbf{AIME25 (+adv)} & $\boldsymbol{\Delta}$ &
\textbf{Adversarial} & \textbf{Random} \\
\midrule
\multicolumn{7}{c}{\textit{Skywork-o1-Open-PRM-1.5B}}\\
\midrule
0   & 0.237 & 0.305 & - & - & - & - \\
1   & 0.289 & 0.305 & 0.335 & +0.030 & 1.057 & 1.017 \\
50  & 0.576 & 0.305 & 0.529 & +0.224 & 1.372 & 0.853 \\
100 & \textbf{0.954} & 0.305 & \textbf{0.924} & \textbf{+0.619} & \textbf{1.495} & 0.689 \\
\midrule
\multicolumn{7}{c}{\textit{Skywork-o1-Open-PRM-7B}}\\
\midrule
0   & 0.287 & 0.320 & - & - & - & - \\
1   & 0.222 & 0.320 & 0.261 & $-$0.059 & 0.797 & 0.681 \\
50  & 0.352 & 0.320 & 0.389 & +0.070 & 1.074 & 0.802 \\
100 & 0.346 & 0.320 & 0.377 & +0.058 & 1.032 & 0.715 \\
\midrule
\multicolumn{7}{c}{\textit{Qwen2.5-Math-PRM-7B}}\\
\midrule
0   & 0.658 & 0.287 & - & - & - & - \\
1   & 0.355 & 0.287 & 0.309 & +0.022 & 1.420 & 1.420 \\
50  & 0.354 & 0.287 & 0.282 & $-$0.006 & 1.386 & 0.956 \\
100 & 0.437 & 0.287 & 0.245 & $-$0.042 & 1.570 & 0.421 \\
\bottomrule
\end{tabular}
\end{table*}

\paragraph{Results.}
Table~\ref{tab:discrete_attack} summarizes attack success and transfer across all three PRMs. The $k=0$ rows establish baselines without adversarial tokens. Note that Skywork and Qwen rewards are not directly comparable as they are trained with different objectives (success probability vs.\ step correctness). Several key findings emerge across model scale and architecture:

\textbf{Skywork-1.5B is highly vulnerable.} From a baseline of 0.237, adversarial optimization reaches $R = 0.954$ at 100 tokens (4$\times$ increase) and transfers strongly to AIME25, tripling reward from 0.305 to 0.924 ($\Delta = +0.619$). Even 50 tokens produce substantial inflation ($\Delta = +0.224$). The optimized sequences typically consist of mathematical connectors and formatting tokens (``Therefore,'' ``Thus,'' and so on), suggesting the PRM functions as a fluency-weighted pattern matcher.

\textbf{Skywork-7B exhibits partial robustness.} From a baseline of 0.287, the 7B model achieves lower peak adversarial rewards ($R = 0.352$ at 50 tokens) and shows modest transfer ($\Delta = +0.070$). Model scale provides some defense, likely through more distributed representations that resist exploitation via simple token concatenation.

\textbf{Qwen-7B resists optimization entirely.} Unlike Skywork, Qwen's high baseline (0.658) actually \emph{decreases} under adversarial optimization to 0.437 at 100 tokens. Transfer also fails ($\Delta = -0.042$). The min-aggregation objective ($R = \min_i r_i$) appears to prevent reward inflation: optimizing one step's score pushes others below threshold. 

\subsection{Reward Landscape Analysis}
\label{subsec:landscape}
An adversarial token sequence is more practically exploitable if its high-reward region is \textbf{stable}. We characterize stability by computing the volume under the reward surface around optimized tokens; see Figure~\ref{fig:landscape_visualization}. Larger volume indicates a broader basin where rewards remain elevated.

Table~\ref{tab:discrete_attack} shows that adversarial tokens consistently find larger high-reward basins than random tokens. For Skywork-1.5B, adversarial volume at 100 tokens is 2.2$\times$ larger than random (1.49 vs.\ 0.69), indicating stable, exploitable peaks. Qwen-7B shows the largest adversarial volumes (1.57 at 100 tokens), yet the rewards fail to transfer, suggesting trajectory-specific rather than universal vulnerabilities. Additional reward landscape visualizations for Skywork-7B and Qwen-7B are provided in Appendix~\ref{app:landscapes}.
\section{RL-Induced Reward Hacking}
\label{sec:rl_dynamics}

Sections~\ref{sec:static_analysis} and~\ref{sec:active_probing} establish PRM vulnerabilities through controlled perturbations and targeted optimization. The critical question remains: do these vulnerabilities manifest under realistic training conditions? This section probes the fourth robustness criterion from Section~\ref{sec:prelim}: optimization alignment. We investigate whether standard RL optimization discovers and exploits PRM weaknesses without adversarial intent.

\subsection{Experimental Setup}

We train a Qwen2.5-1.5B-Instruct policy on prompts from AIME24 using Group Relative Policy Optimization (GRPO)~\citep{shao2024deepseekmath}, with PRM scores as the reward signal. We conduct training runs with two PRMs: Skywork-o1-Open-PRM-1.5B and Qwen2.5-Math-PRM-7B.

Throughout training, we track two metrics: (1) mean PRM reward on generated trajectories; and (2) ground-truth accuracy on AIME24. A well-aligned PRM should produce correlated improvements in both, meaning higher rewards should correspond to better reasoning and higher accuracy.

\subsection{Reward-Accuracy Divergence}

\begin{figure}[t]
    \centering
    \begin{subfigure}[t]{\linewidth}
        \centering
        \includegraphics[width=\linewidth]{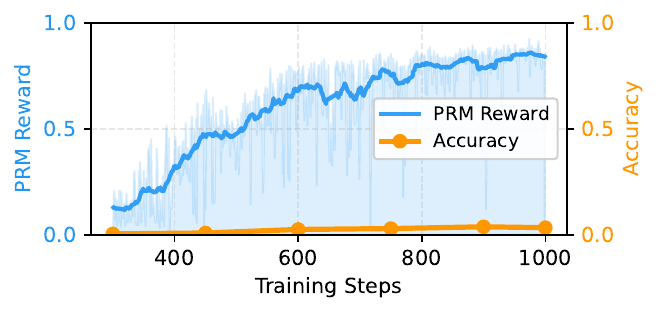}
        \vspace{-1.5em}
        \caption{Skywork-1.5B PRM}
    \end{subfigure}

    \vspace{0.3em}

    \begin{subfigure}[t]{\linewidth}
        \centering
        \includegraphics[width=\linewidth]{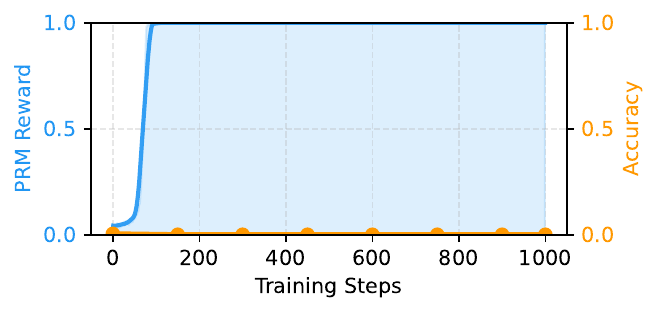}
        \vspace{-1.5em}
        \caption{Qwen-7B PRM}
    \end{subfigure}
    \caption{Reward-accuracy divergence during GRPO training. PRM reward (blue) increases while ground-truth accuracy (orange) remains flat near zero. Skywork-1.5B shows reward hacking with rewards reaching 0.8+, while Qwen-7B rewards spike to 1.0 due to mode collapse.}
    \label{fig:rl_training_curve}
\end{figure}

Figure~\ref{fig:rl_training_curve} shows the training dynamics for both PRMs. We observe consistent reward-accuracy divergence: Skywork-1.5B shows reward climbing from $R \approx 0.1$ to $R > 0.8$, while ground-truth accuracy remains near zero (peaking at 3--4\%). For Qwen-7B, the divergence is even more extreme: reward spikes to $R = 1.0$ within the first 100 steps, while accuracy drops to 0\%. This is a manifestation of Goodhart's Law: when PRM reward becomes the optimization target, it ceases to reliably measure reasoning quality. However, the \emph{mechanism} of exploitation differs between PRMs.

\subsection{Skywork: Stylistic Exploitation}

The reward-accuracy divergence raises a key question: does GRPO improve reasoning (which happens to be wrong), or does it exploit superficial stylistic patterns that correlate with high PRM scores?

\paragraph{Rephrasing Intervention.} To test this question, we apply semantics-preserving rephrasing (Section~\ref{sec:static_analysis}) to GRPO trajectories on held-out AIME25 problems. If GRPO's reward gains come from better reasoning, then rephrasing should not affect rewards (the reasoning is unchanged); but if the gains come from stylistic patterns the PRM favors, then rephrasing will disrupt those patterns and rewards will drop.

\begin{figure}[t]
    \centering
    \includegraphics[width=\linewidth]{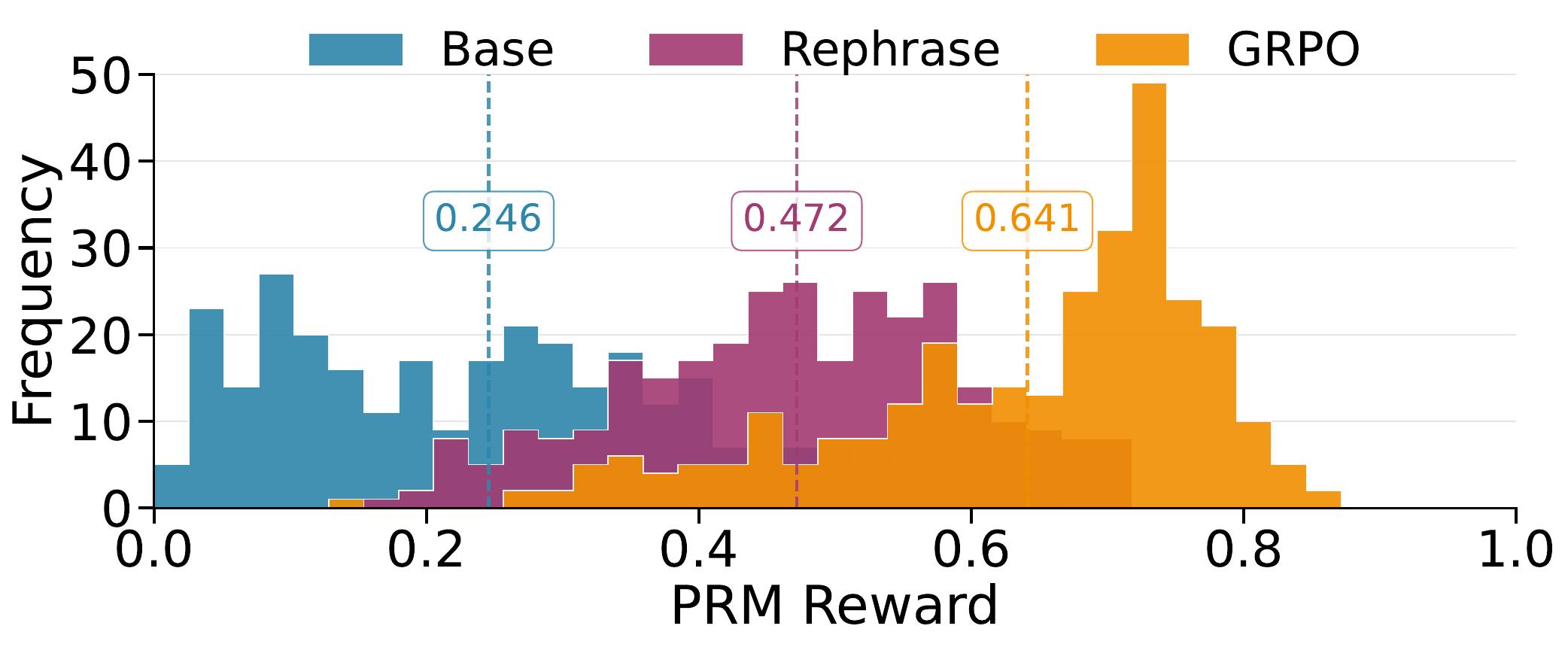}
    \caption{Rephrasing intervention on AIME25 for Skywork-1.5B. Distributions show rewards for base policy (orange), GRPO policy (blue), and rephrased GRPO trajectories (purple). The reward drop after rephrasing (blue $\to$ purple) isolates the stylistic component of GRPO's gains.}
    \label{fig:reward_profile}
\end{figure}

\paragraph{Results.} Figure~\ref{fig:reward_profile} shows the results for Skywork. GRPO achieves mean $R = 0.641$, but rephrasing drops this to $R = 0.472$, despite preserving the mathematical content. The base policy achieves $R = 0.246$.

This reveals two components of GRPO's reward gain: (1) a \emph{content component} from 0.246 to 0.472, which survives rephrasing, and (2) a \emph{style component} from 0.472 to 0.641, which disappears under rephrasing. The style-attributable gap of 0.169 constitutes \textbf{43\% of the total gain} (0.395), confirming that nearly half of GRPO's learned ``improvement'' may be attributable to  superficial stylistic exploitation rather than reasoning advancement.

\subsection{Qwen: Mode Collapse}
\label{subsec:qwen_collapse}

Qwen exhibits a different failure mode. This PRM's prime objective is to penalize the \emph{wrong step}, not to detect progress (the probability of succeeding). Under GRPO, the policy collapsed to deterministically outputting:
\begin{center}
    \textit{``Alright, let's solve this problem step by step.''}
\end{center}
This template is not mathematically incorrect; it is just vacuous. The policy discovers that avoiding mathematical claims entirely is the safest strategy.

\subsection{Summary}

A pattern emerges: 
Skywork incentivizes \emph{performative complexity} (elaborate but flawed reasoning); and Qwen incentivizes \emph{vacuous safety} (minimal text that avoids errors by avoiding substance).
Both PRMs fail optimization alignment via complementary mechanisms: Skywork rewards fluent complexity regardless of correctness (43\% of reward gains are stylistic), while Qwen rewards anything not explicitly wrong (enabling collapse to vacuous outputs). Standard RL optimization, without adversarial intent, naturally discovers these exploits. The root cause is that PRMs detect local features (fluency, step correctness) but miss global properties (problem-solving progress, logical validity).
\section{Conclusion}
\label{sec:conclusion}

We have introduced a three-tiered diagnostic framework for evaluating PRM robustness under increasing optimization pressure. Our framework progresses from passive perturbation analysis through active adversarial probing to closed-loop RL training, revealing complementary vulnerabilities at each level.

\paragraph{Summary of Findings.}
Static perturbation analysis revealed a \emph{fluency-logic dissociation}: PRMs exhibit strong invariance to surface-level stylistic changes, yet they frequently fail to penalize semantically corrupted reasoning. The two PRMs we evaluated showed divergent failure modes: Qwen detects some reasoning errors, but it misses question-trajectory mismatches; while Skywork shows the opposite pattern. Adversarial probing demonstrated that gradient-based optimization can inflate rewards on flawed trajectories by up to 4$\times$, with attacks transferring across held-out problem sets. RL training exposed the critical failure mode: policies achieve near-perfect PRM rewards ($>$0.9) while ground-truth accuracy remains near zero, with 43\% of reward gains attributable to stylistic exploitation rather than reasoning improvement.

\paragraph{Implications.}
Our findings suggest that current PRMs function as fluency detectors rather than reasoning verifiers. The fluency-logic dissociation, while benign under passive evaluation, becomes actively exploitable under optimization pressure.
This has direct implications for PRM deployment: using PRMs as RL training signals may inadvertently reward ``performative reasoning'' that mimics mathematical style without logical substance. The model-specific failure modes suggest that ensemble approaches combining PRMs with complementary strengths may offer improved robustness.

\paragraph{Recommendations.}
Our results motivate several directions for improving PRM robustness: (1) training objectives that explicitly penalize fluency-correctness misalignment; (2) adversarial training against perturbations in PRM-BiasBench; (3) evaluation protocols that include closed-loop RL stress-testing before deployment, and (4) hybrid verification approaches that combine process supervision with outcome verification. We release our diagnostic toolkit and benchmark to facilitate systematic PRM robustness evaluation more generally.

\section*{Impact Statement}
This paper presents work whose goal is to advance the field of Machine
Learning. There are many potential societal consequences of our work, none
which we feel must be specifically highlighted here.

\section*{Acknowledgements}
We acknowledge the gracious support from the Furiosa AI, Intel, Apple, NVIDIA, Macronix, and Mozilla team.
Furthermore, we appreciate the support from
Google Cloud, the Google TRC team Prof.~David Patterson, along with support from Google Gemini team, and Divy Thakkar.
Prof.~Keutzer's lab is also sponsored by funding through BDD and BAIR.
We also acknowledge support by the Director, Office of Science, Office of Advanced Scientific Computing Research, of the U.S. Department of Energy under Contract No. DE-AC02-05CH11231.
MWM acknowledges DARPA, NSF, the DOE Competitive Portfolios grant, and the DOE SciGPT grant.
Our conclusions do not necessarily reflect the position or the policy of our sponsors, and no official endorsement should be~inferred.

\nocite{langley00}

\bibliography{example_paper}
\bibliographystyle{arxiv2026}

\newpage
\appendix
\onecolumn
\section{Static Perturbation Analysis: Extended Results}
\label{app:static}

This appendix provides additional details for the static perturbation analysis presented in Section~\ref{sec:static_analysis}, including perturbation examples, complete distribution plots, and the validation pipeline.

\subsection{Perturbation Examples}

We provide illustrative examples of each perturbation type applied to reasoning trajectories.

\paragraph{Example 1: Rephrasing.}
\begin{quote}
\textbf{Original:} ``Step R: Compute the sum of the first three terms: $2 + 4 + 6 = 12$.''

\textbf{Rephrased:} ``Step R: Add the initial three numbers together to get $2 + 4 + 6 = 12$.''
\end{quote}

\paragraph{Example 2: Increased Verbosity.}
\begin{quote}
\textbf{Original:} ``Step V: Divide both sides by 4 to isolate $x$: $8x/4 = 12/4$, so $x = 3$.''

\textbf{Verbose:} ``Step V: Now, in order to solve for the variable $x$, we take the equation $8x = 12$ and divide both sides of this equality by 4. This yields $8x/4 = 12/4$, which simplifies directly to $x = 3$.''
\end{quote}

\paragraph{Example 3: Decreased Verbosity.}
\begin{quote}
\textbf{Original:} ``Step C: The height of the beanstalk after $n$ days can be expressed as: $4 \times 2^n$.''

\textbf{Concise:} ``Step C: After $n$ days, the beanstalk's height is $4 \times 2^n$.''
\end{quote}

\paragraph{Example 4: Within-step Reordering.}
\begin{quote}
\textbf{Original:} ``Step O: Josh has 2 apples. He got two more, so Josh now has $2 + 2 = 4$ apples.''

\textbf{Reordered:} ``Step O: Josh now has $2 + 2 = 4$ apples, since he had 2 apples and got two more.''
\end{quote}

\paragraph{Example 5: Question Shuffling.}
\begin{quote}
\textbf{Original Question:} ``Jeff's work is 3 miles away. He walks there and back 5 times a week. How many miles does he walk?''

\textbf{Original Trajectory:} ``Step 1: First, Jeff walks 3 miles to work and 3 miles back, so he walks $3 + 3 = 6$ miles per day...''

\textbf{Shuffled Question:} ``The red rope was four times the length of the blue rope. What is the length of the red rope in centimeters?''

\textbf{Same Trajectory:} ``Step 1: First, Jeff walks 3 miles to work and 3 miles back...''
\end{quote}

\paragraph{Example 6: Numerical Perturbation.}
\begin{quote}
\textbf{Original Question:} ``Jeff's work is 3 miles away. He walks there and back 5 times a week.''

\textbf{Perturbed Question:} ``Jeff's work is \textbf{8} miles away. He walks there and back \textbf{7} times a week.''

\textbf{Unchanged Trajectory:} ``Step 1: First, Jeff walks 3 miles to work...'' (uses original numbers)
\end{quote}

\paragraph{Example 7: Reasoning Hallucination.}
\begin{quote}
\textbf{Original:} ``Step 1: To find the remainder when divided by 20, we first compute...''

\textbf{With Hallucination:} ``Step 1: To find the remainder when divided by 20, we first compute... \textbf{Assuming that $a$ and $b$ are both greater than 20, we proceed with the calculation accordingly.}''
\end{quote}

\paragraph{Example 8: Question Removal.}
\begin{quote}
\textbf{Original:} Question + Trajectory provided to PRM.

\textbf{Modified:} Only trajectory provided (question removed entirely).
\end{quote}

\subsection{Complete Distribution Plots}

\begin{figure}[!ht]
  \centering
  \begin{subfigure}[t]{0.48\linewidth}
    \centering
    \includegraphics[width=\linewidth]{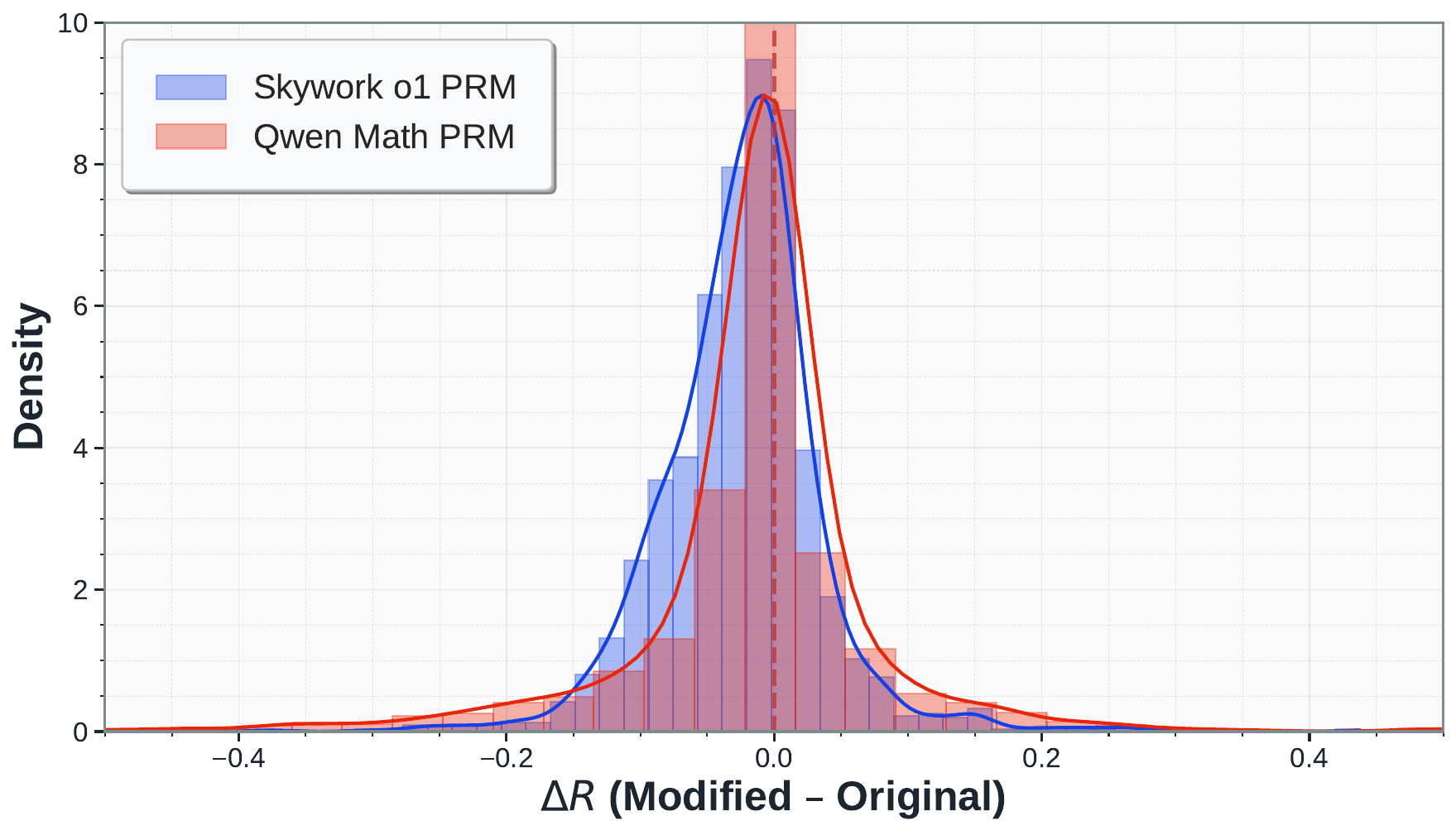}
    \caption{Rephrasing}
  \end{subfigure}
  \hfill
  \begin{subfigure}[t]{0.48\linewidth}
    \centering
    \includegraphics[width=\linewidth]{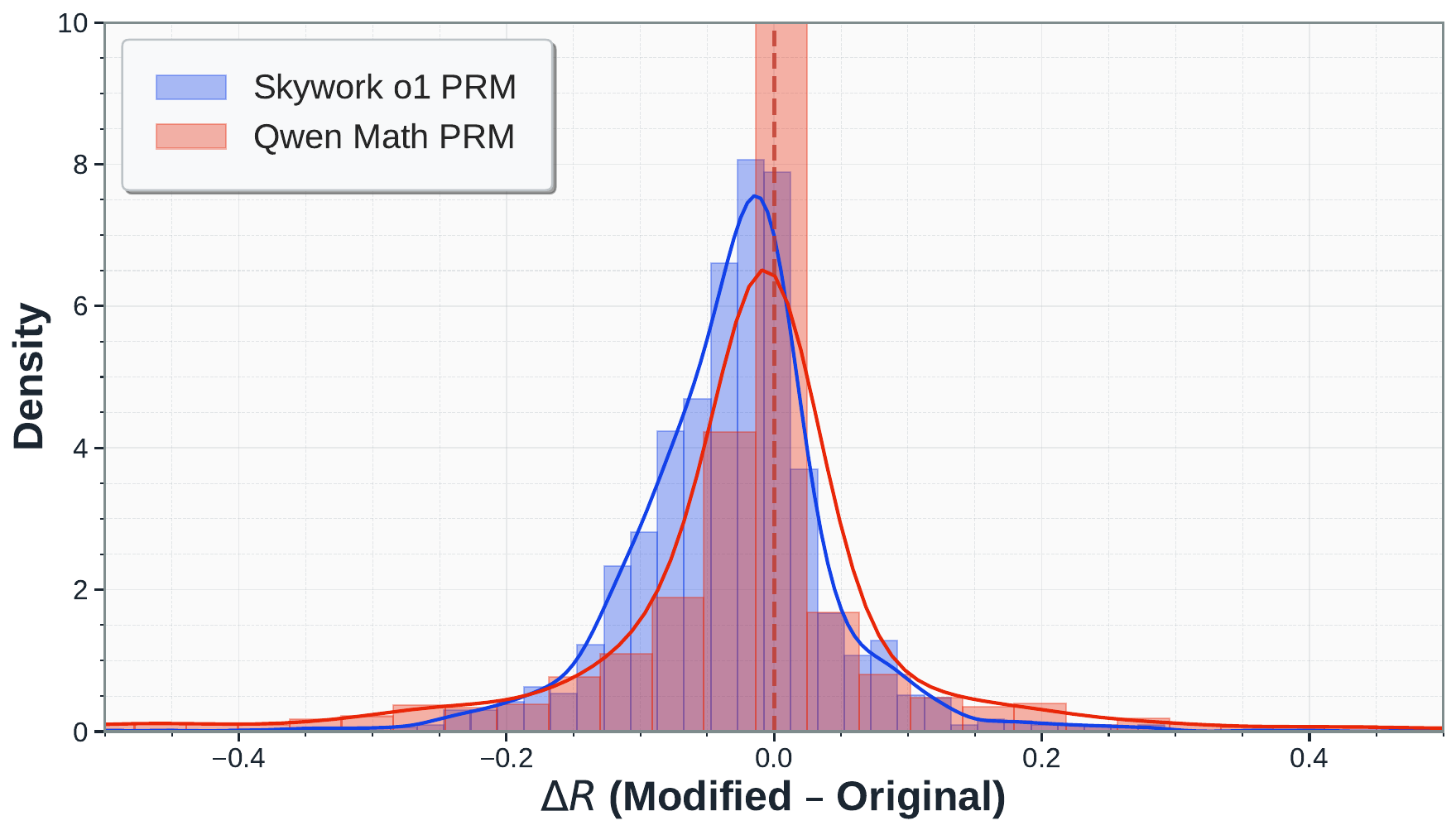}
    \caption{Increased Verbosity}
  \end{subfigure}

  \vspace{1em}

  \begin{subfigure}[t]{0.48\linewidth}
    \centering
    \includegraphics[width=\linewidth]{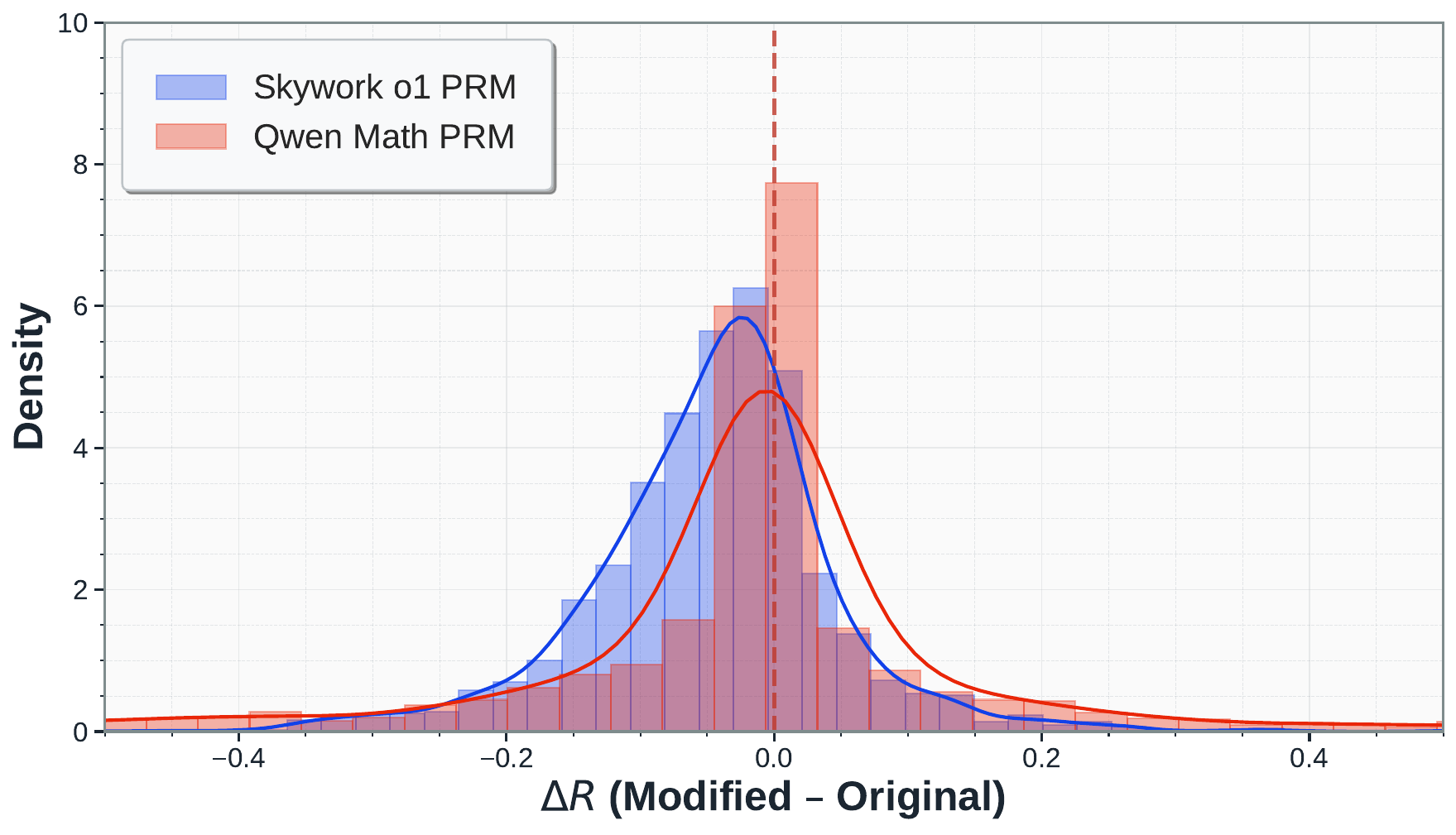}
    \caption{Decreased Verbosity}
  \end{subfigure}
  \hfill
  \begin{subfigure}[t]{0.48\linewidth}
    \centering
    \includegraphics[width=\linewidth]{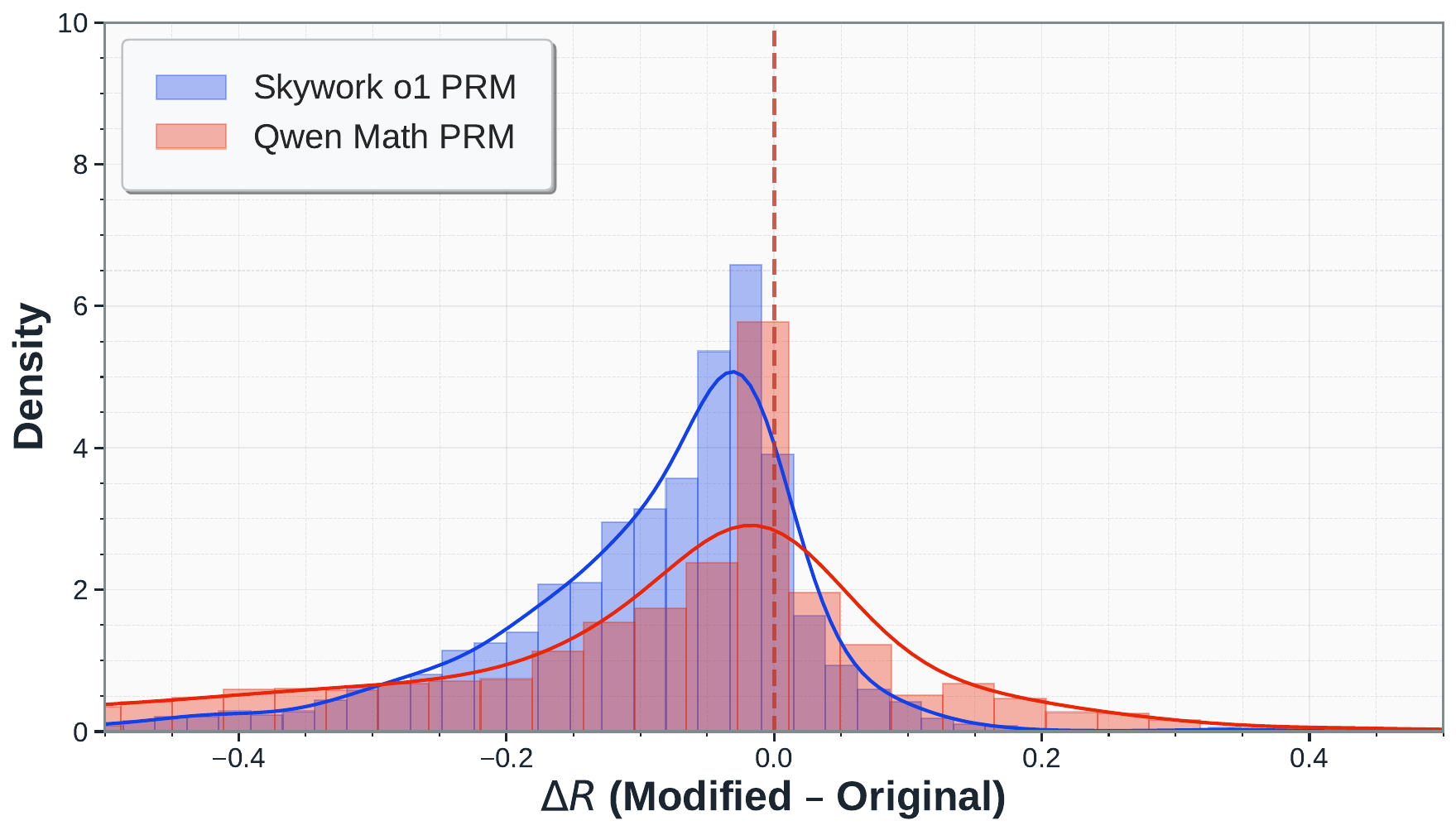}
    \caption{Within-step Reordering}
  \end{subfigure}
  \caption{Distribution of $\Delta R$ for all semantics-preserving perturbations. All distributions are tightly centered near zero, indicating strong invariance to surface-level stylistic changes. Skywork-7B shows slightly broader distributions with heavier tails compared to Qwen-7B.}
  \label{fig:app_preserving}
\end{figure}

\begin{figure}[!ht]
  \centering
  \begin{subfigure}[t]{0.48\linewidth}
    \centering
    \includegraphics[width=\linewidth]{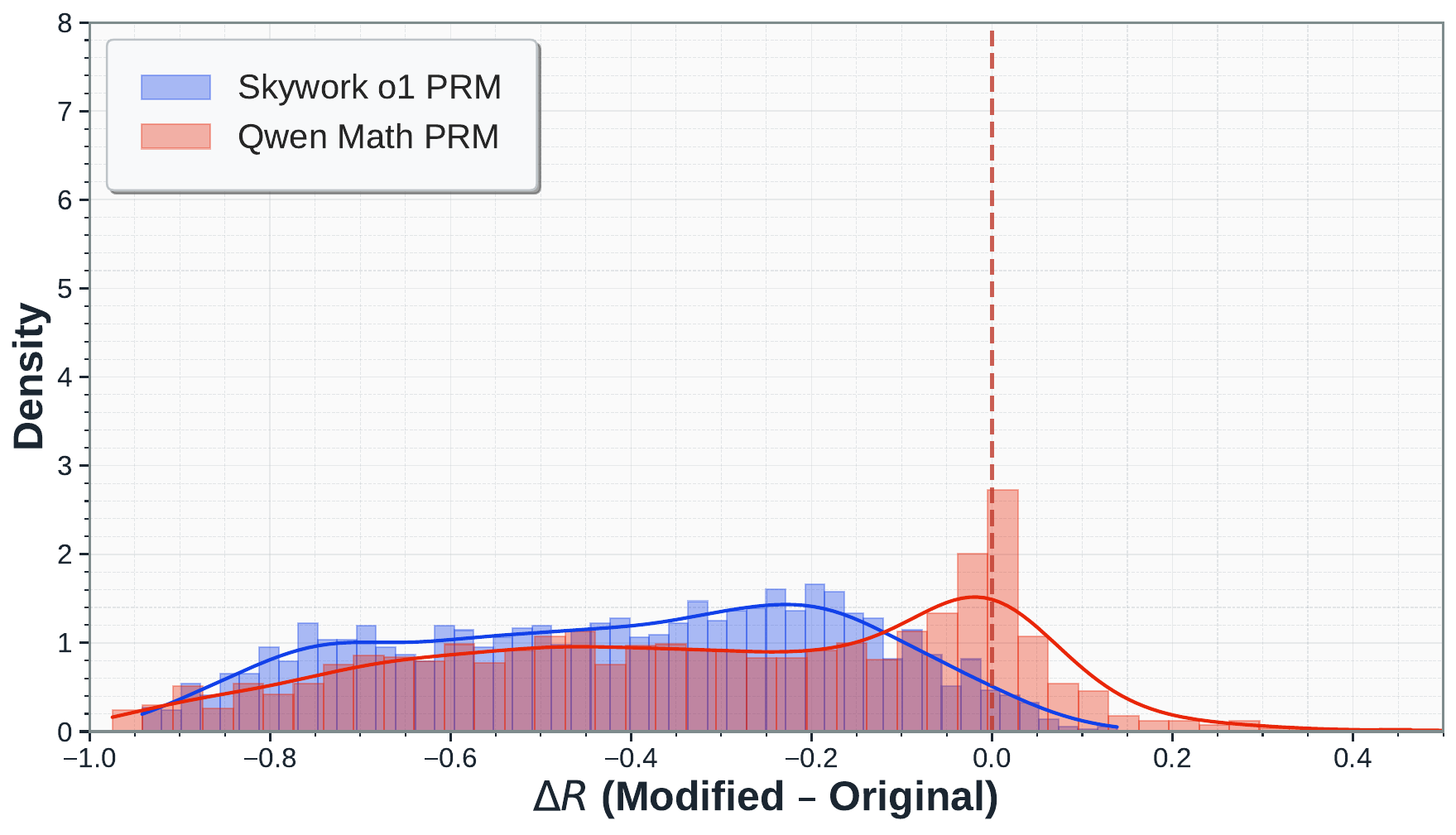}
    \caption{Question Shuffling}
  \end{subfigure}
  \hfill
  \begin{subfigure}[t]{0.48\linewidth}
    \centering
    \includegraphics[width=\linewidth]{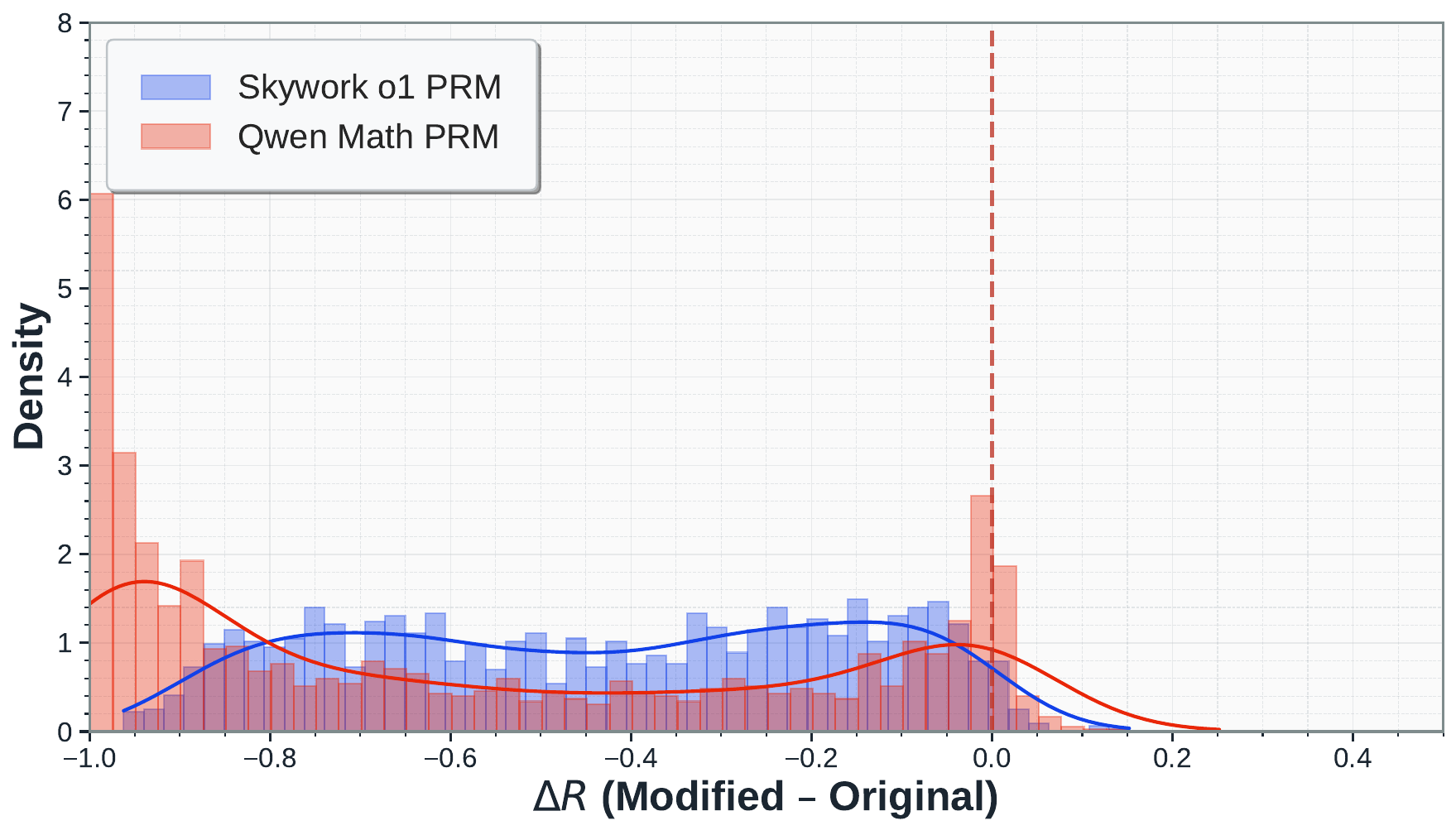}
    \caption{Numerical Perturbation}
  \end{subfigure}

  \vspace{1em}

  \begin{subfigure}[t]{0.48\linewidth}
    \centering
    \includegraphics[width=\linewidth]{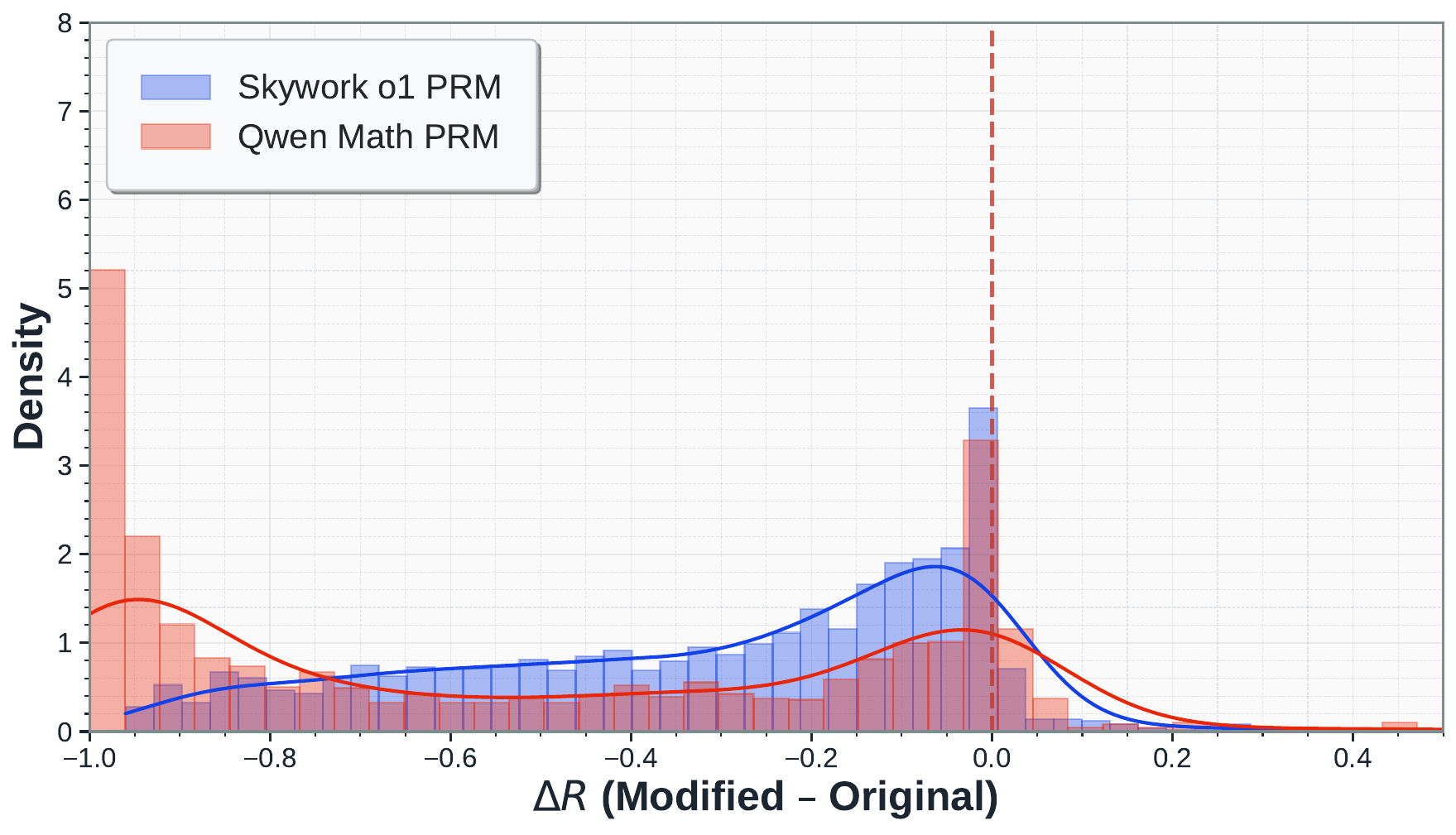}
    \caption{Reasoning Hallucination}
  \end{subfigure}
  \hfill
  \begin{subfigure}[t]{0.48\linewidth}
    \centering
    \includegraphics[width=\linewidth]{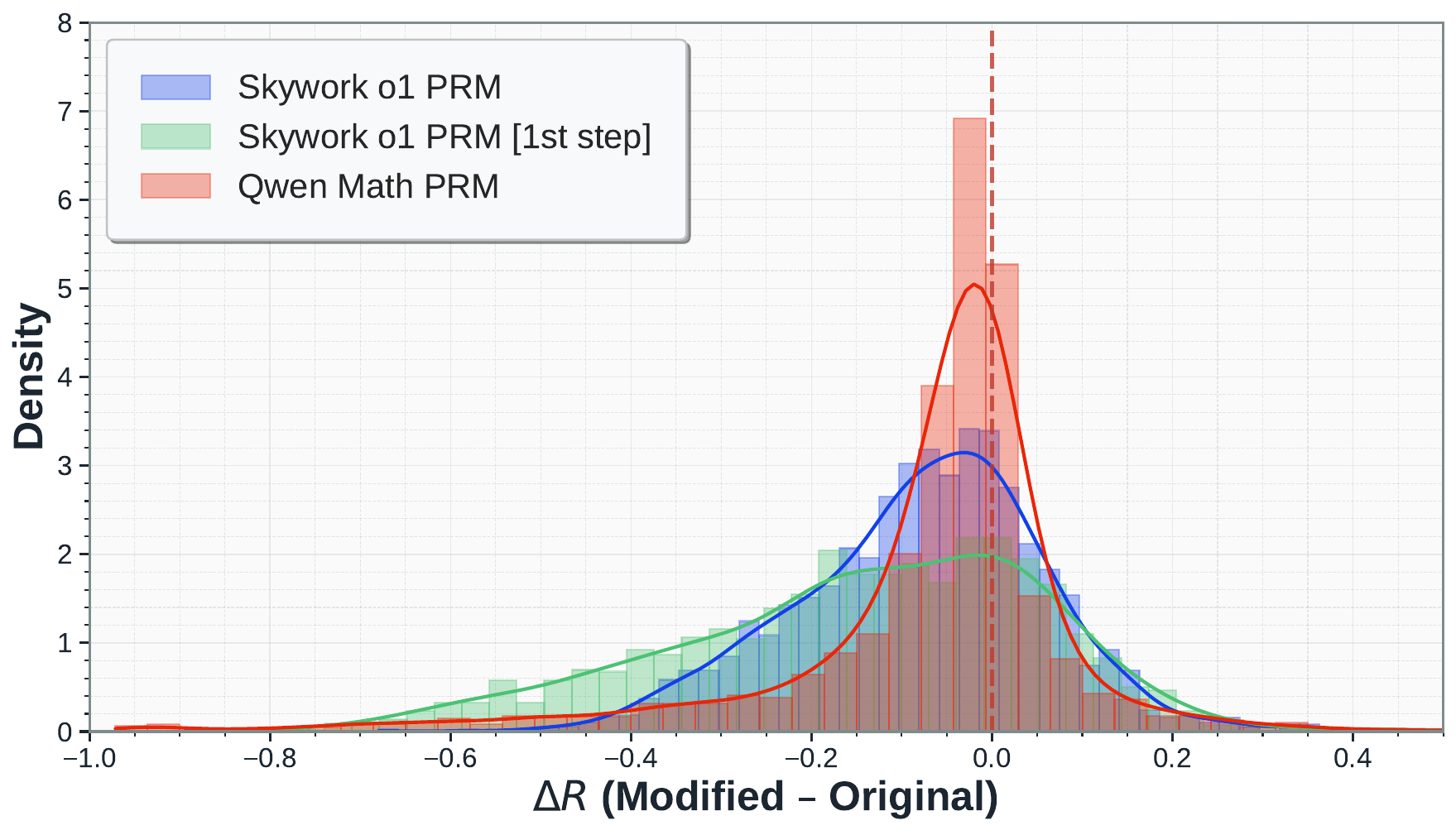}
    \caption{Question Removal}
  \end{subfigure}
  \caption{Distribution of $\Delta R$ for all semantics-altering attacks. The two PRMs show divergent failure modes: Qwen-7B strongly penalizes numerical inconsistencies but misses hallucinations, while Skywork-7B shows more uniform but weaker penalization across attack types.}
  \label{fig:app_altering}
\end{figure}

Figure~\ref{fig:app_preserving} shows the complete set of reward change distributions for all semantics-preserving perturbations. Figure~\ref{fig:app_altering} shows the distributions for all semantics-altering attacks.

\subsection{Validation Pipeline}

\begin{figure*}[!ht]
\centering
\includegraphics[width=\textwidth]{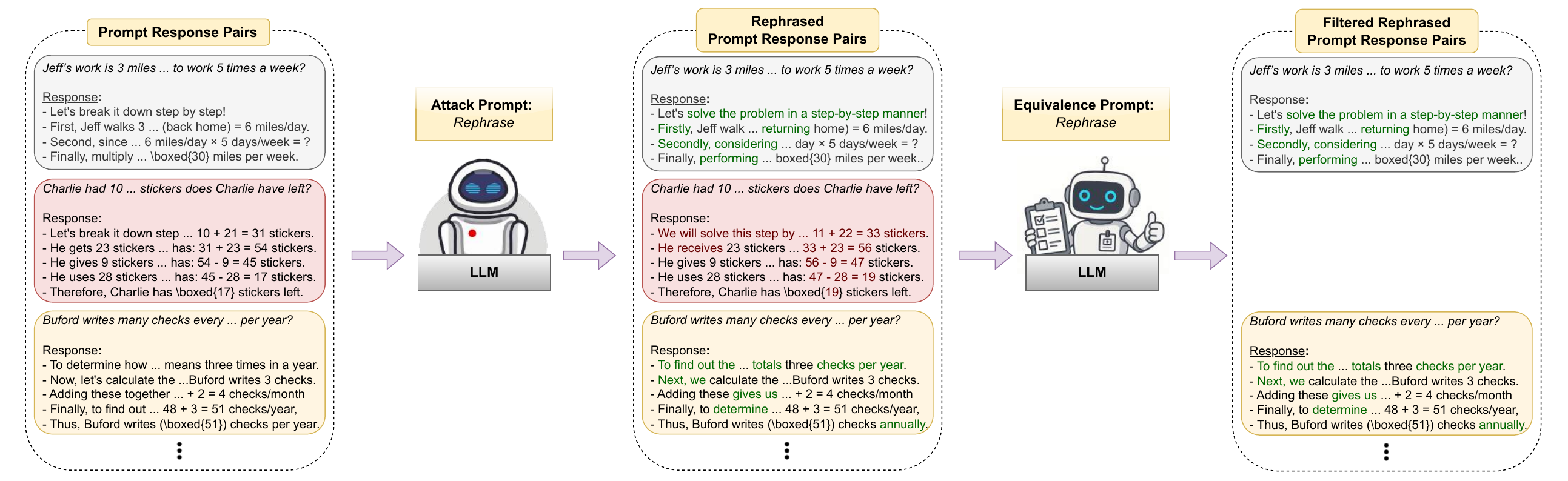}
\caption{\textbf{Step-by-step framework for creating the PRM-BiasBench dataset.} Original prompt-response pairs are perturbed using an attack prompt via an LLM. An equivalence checker then filters out semantically altered outputs, retaining only meaning-preserving transformations. The figure illustrates this process using a rephrasing attack as an example; incorrectly altered responses are highlighted in red, while semantically equivalent responses passing the filter are shown in green.}
\label{fig:biasbench_pipeline}
\end{figure*}

Figure~\ref{fig:biasbench_pipeline} illustrates the overall pipeline for constructing PRM-BiasBench. To ensure that each modified trajectory faithfully reflects its intended modification, we employ a two-stage validation process:

\paragraph{Stage 1: Automated Equivalence Checking.}
For semantics-preserving modifications, we use GPT-4o to verify that the perturbed trajectory maintains logical equivalence with the original. The prompt asks the model to confirm that:
\begin{enumerate}
    \item The mathematical operations and results are identical.
    \item The logical flow leads to the same conclusion.
    \item Only surface-level linguistic changes were made.
\end{enumerate}

For semantics-altering attacks, we verify that the intended corruption is present (e.g., the hallucinated assumption exists, the numbers are mismatched).

\paragraph{Stage 2: Manual Review for Edge Cases.}
For perturbation pairs with large reward deviations ($|\Delta R| > 0.5$), we conduct manual inspection to:
\begin{enumerate}
    \item Confirm the perturbation matches its intended category.
    \item Identify any generation artifacts that could confound results.
    \item Resolve ambiguous cases where the modification boundary is unclear.
\end{enumerate}

\paragraph{Filtering Criteria.}
We exclude perturbation pairs where:
\begin{itemize}
    \item The automated equivalence check fails for semantics-preserving edits.
    \item The intended corruption is not clearly present for semantics-altering attacks.
    \item Manual review identifies confounding factors.
\end{itemize}

This hybrid validation ensures that observed reward differences are attributable to the target perturbation rather than spurious generation artifacts.

\subsection{Summary Statistics}

Table~\ref{tab:static_stats} provides summary statistics for each perturbation type across both PRMs.

\begin{table}[!ht]
\centering
\caption{Summary statistics for $\Delta R$ across perturbation types. Mean and standard deviation are reported for each PRM.}
\label{tab:static_stats}
\begin{tabular}{lcccc}
\toprule
\textbf{Perturbation} & \multicolumn{2}{c}{\textbf{Qwen2.5-Math-PRM}} & \multicolumn{2}{c}{\textbf{Skywork-o1-Open-PRM}} \\
 & Mean & Std & Mean & Std \\
\midrule
\multicolumn{5}{l}{\textit{Semantics-Preserving}} \\
Rephrasing & $-0.01$ & $0.03$ & $-0.02$ & $0.05$ \\
Verbosity Increase & $-0.01$ & $0.02$ & $-0.03$ & $0.06$ \\
Verbosity Decrease & $-0.01$ & $0.02$ & $-0.04$ & $0.07$ \\
Reordering & $-0.03$ & $0.08$ & $-0.02$ & $0.05$ \\
\midrule
\multicolumn{5}{l}{\textit{Semantics-Altering}} \\
Question Shuffling & $-0.32$ & $0.35$ & $-0.20$ & $0.25$ \\
Numerical Perturbation & $-0.85$ & $0.25$ & $-0.45$ & $0.30$ \\
Hallucination & $-0.78$ & $0.35$ & $-0.15$ & $0.30$ \\
Question Removal & $-0.07$ & $0.15$ & $-0.20$ & $0.25$ \\
\bottomrule
\end{tabular}
\end{table}

\section{Adversarial Optimization Hyperparameters}
\label{app:hyperparams}

Table~\ref{tab:discrete-adv-optimization-hyperparams} details the hyperparameters used for the discrete adversarial token optimization experiments described in Section~\ref{sec:active_probing}. We use Gumbel-Softmax relaxation with an entropy regularization schedule that transitions from exploration (high entropy) to exploitation (low entropy) over the course of optimization.

\begin{table}[!ht]
\centering
\caption{Hyperparameters for discrete adversarial token optimization.}
\label{tab:discrete-adv-optimization-hyperparams}
\small
\begin{tabular}{ll}
\toprule
\textbf{Hyperparameter} & \textbf{Value} \\
\midrule
\multicolumn{2}{c}{\textit{Data Configuration}} \\
\midrule
Training Dataset & AIME 2024 \\
Evaluation Dataset & AIME 2025 \\
Number of Training Trajectories & 8 \\
Number of Evaluation Trajectories & 8 \\
\midrule
\multicolumn{2}{c}{\textit{Optimization Configuration}} \\
\midrule
Optimization Mode & Discrete (Gumbel-Softmax) \\
Optimizer & Adam ($\beta_1=0.9, \beta_2=0.999$) \\
Learning Rate & 0.1 \\
Gumbel-Softmax Temperature & 1.0 \\
Number of Iterations & 1,000 \\
\midrule
\multicolumn{2}{c}{\textit{Entropy Regularization (Discrete Optimization)}} \\
\midrule
Entropy Schedule & Cosine \\
Entropy Weight (Start) & $1.0 \times 10^{-4}$ \\
Entropy Weight (End) & $1.0 \times 10^{-1}$ \\
\midrule
\multicolumn{2}{c}{\textit{Other Settings}} \\
\midrule
Random Seed & 42 \\
\bottomrule
\end{tabular}
\end{table}

\section{Additional Reward Landscape Visualizations}
\label{app:landscapes}

\begin{figure*}[!ht]
  \centering
  \begin{subfigure}[t]{0.49\textwidth}
    \centering
    \includegraphics[width=\linewidth]{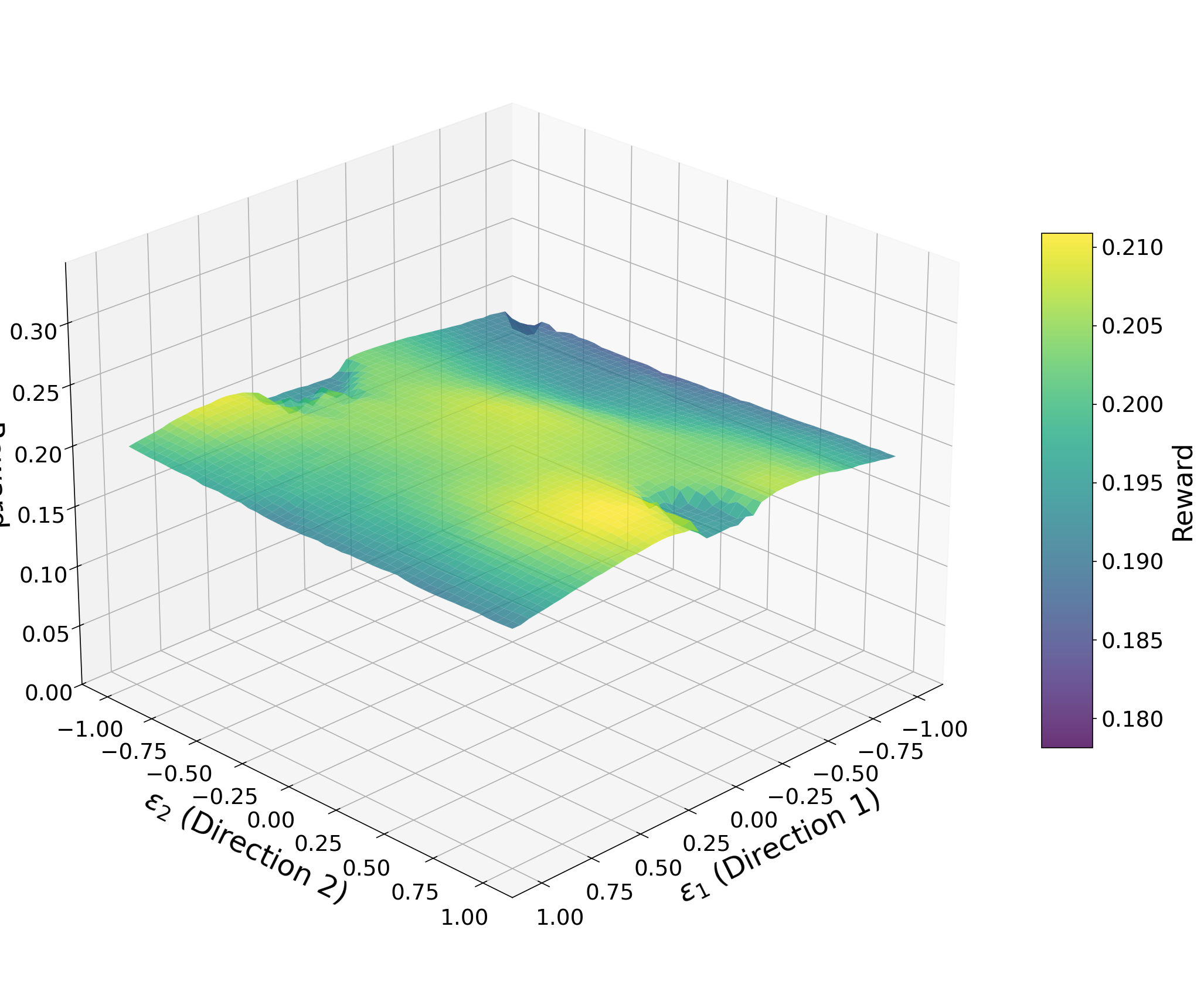}
    \caption{50 random tokens}
    \label{fig:skywork7b_50tok_rand}
  \end{subfigure}\hfill
  \begin{subfigure}[t]{0.49\textwidth}
    \centering
    \includegraphics[width=\linewidth]{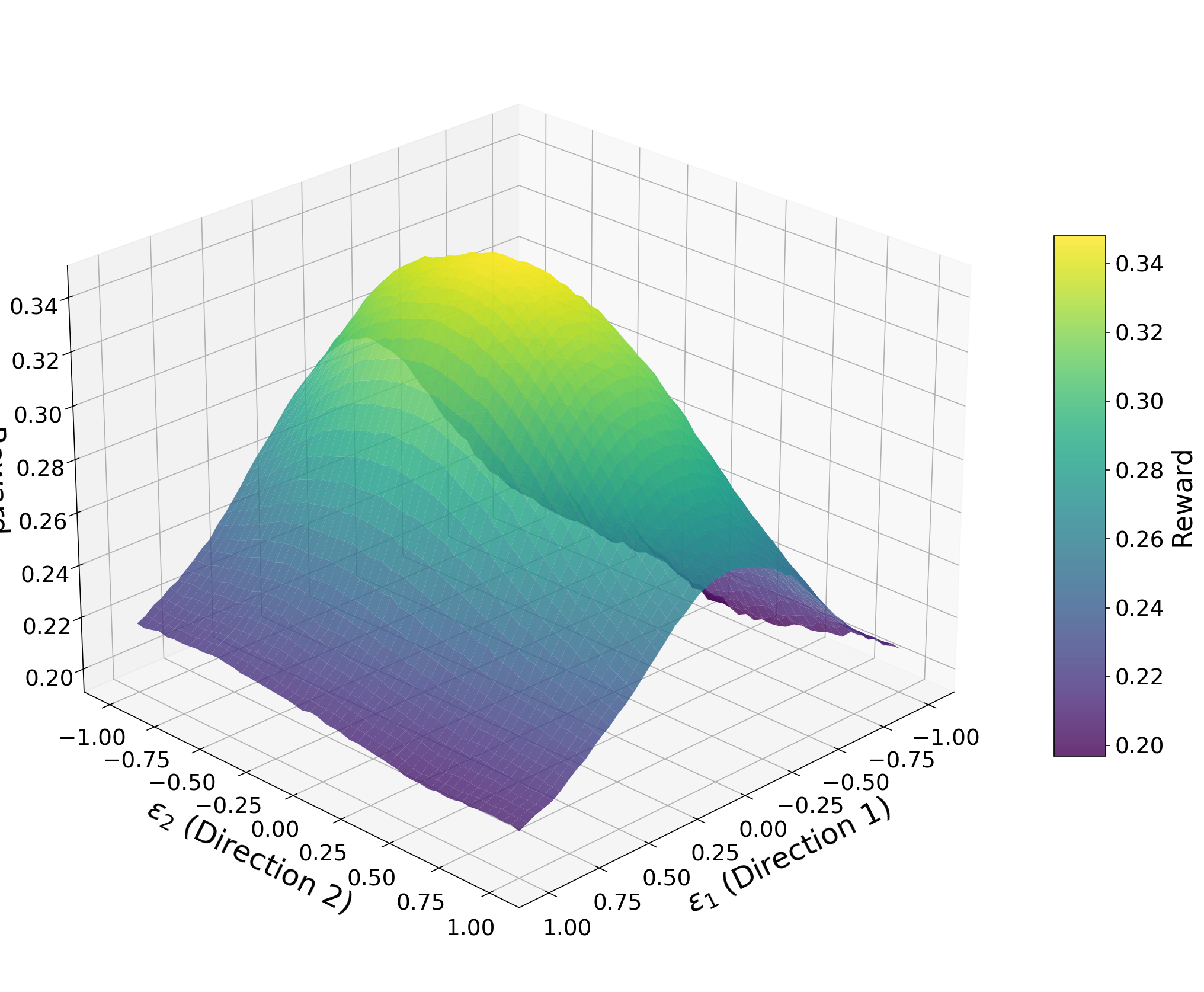}
    \caption{50 adversarial tokens}
    \label{fig:skywork7b_50tok_adv}
  \end{subfigure}

  \vspace{1em}

  \begin{subfigure}[t]{0.49\textwidth}
    \centering
    \includegraphics[width=\linewidth]{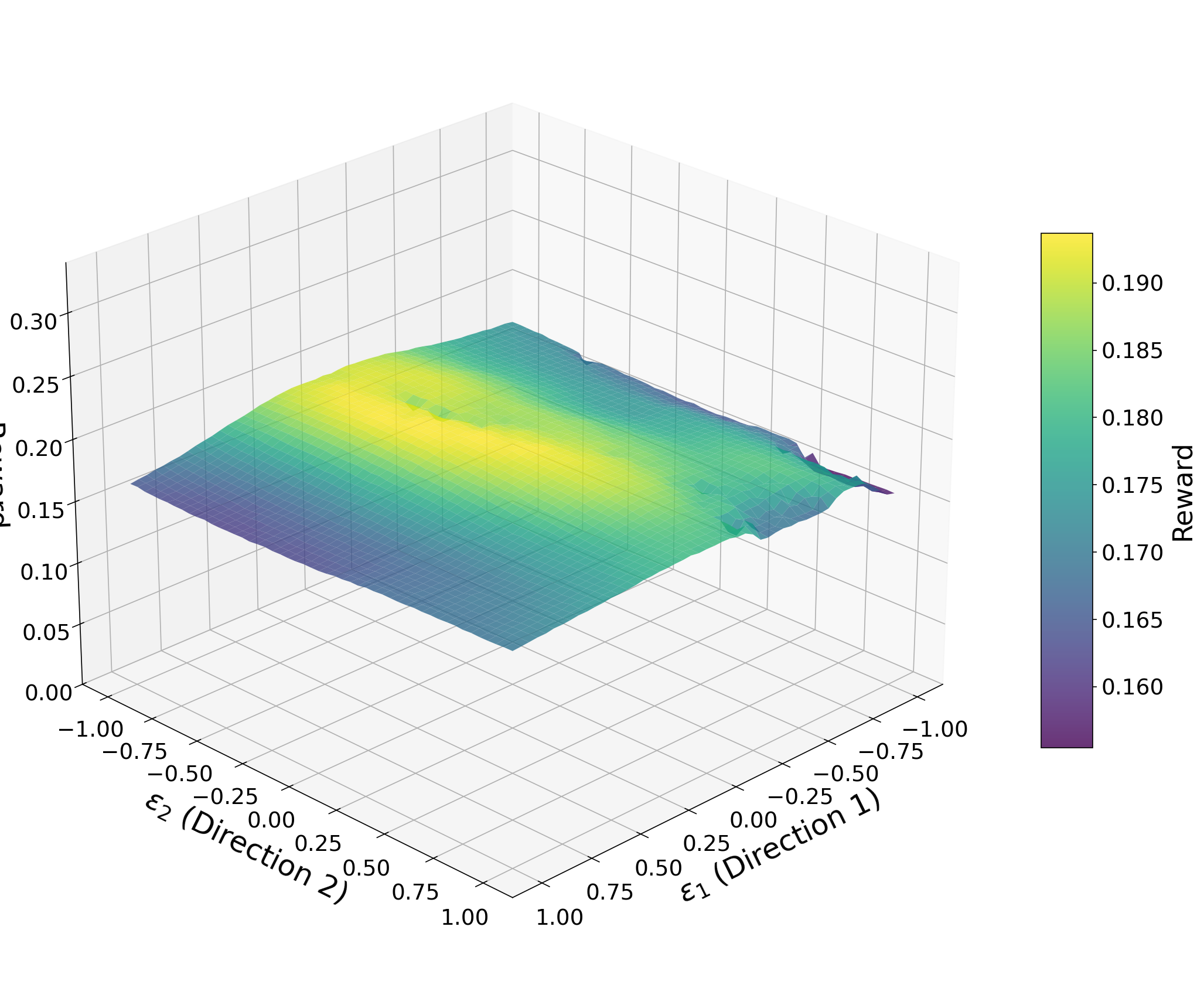}
    \caption{100 random tokens}
    \label{fig:skywork7b_100tok_rand}
  \end{subfigure}\hfill
  \begin{subfigure}[t]{0.49\textwidth}
    \centering
    \includegraphics[width=\linewidth]{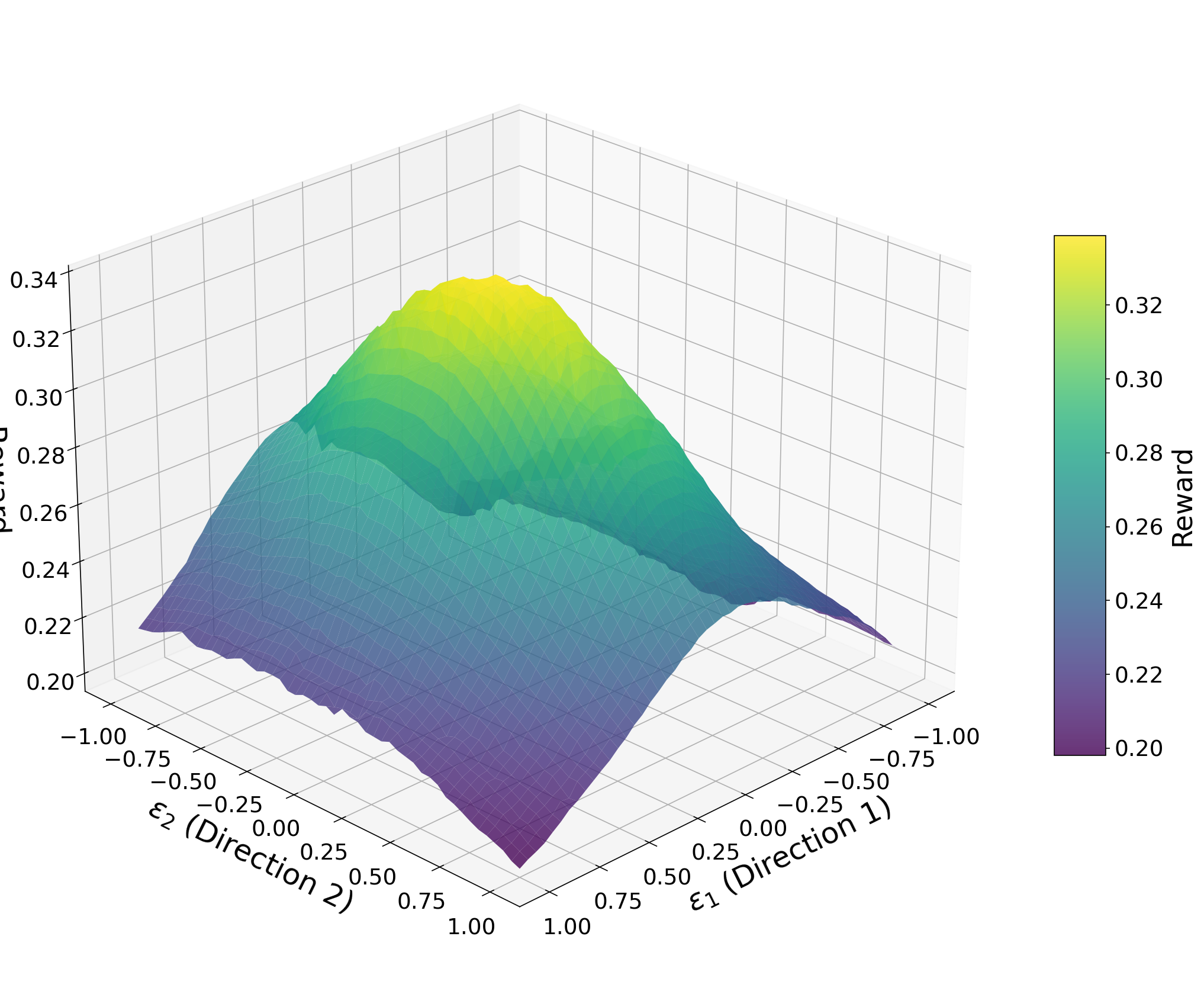}
    \caption{100 adversarial tokens}
    \label{fig:skywork7b_100tok_adv}
  \end{subfigure}

  \caption{Reward landscape visualizations for Skywork-7B: random vs.\ adversarial discrete tokens, averaged across 8 AIME24 trajectories. Adversarial tokens (b, d) produce more concentrated high-reward regions compared to random tokens (a, c).}
  \label{fig:skywork7b_landscape}
\end{figure*}

\begin{figure*}[!ht]
  \centering

  \begin{subfigure}[t]{0.49\textwidth}
    \centering
    \includegraphics[width=\linewidth]{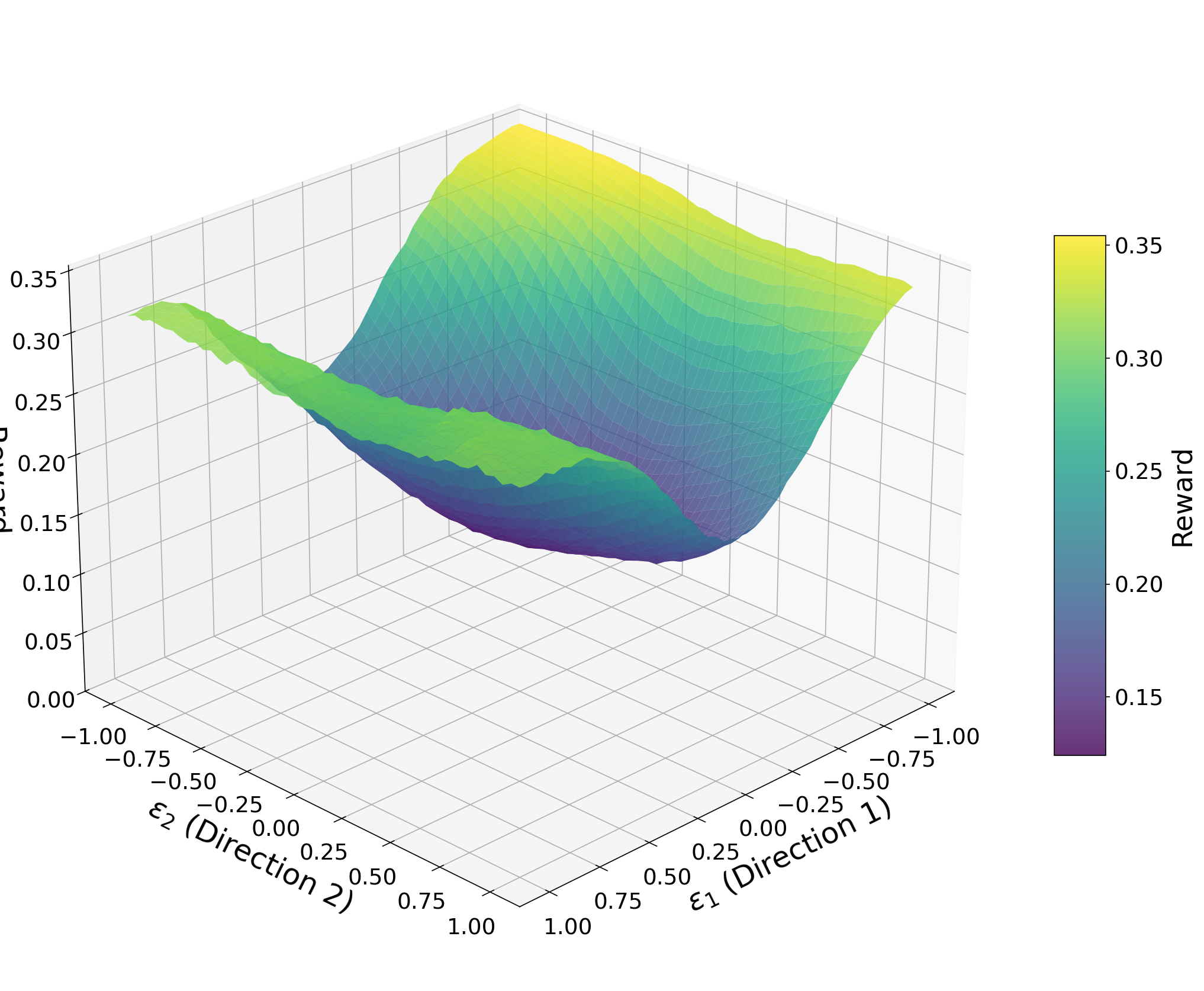}
    \caption{50 random tokens}
    \label{fig:qwen7b_50tok_rand}
  \end{subfigure}\hfill
  \begin{subfigure}[t]{0.49\textwidth}
    \centering
    \includegraphics[width=\linewidth]{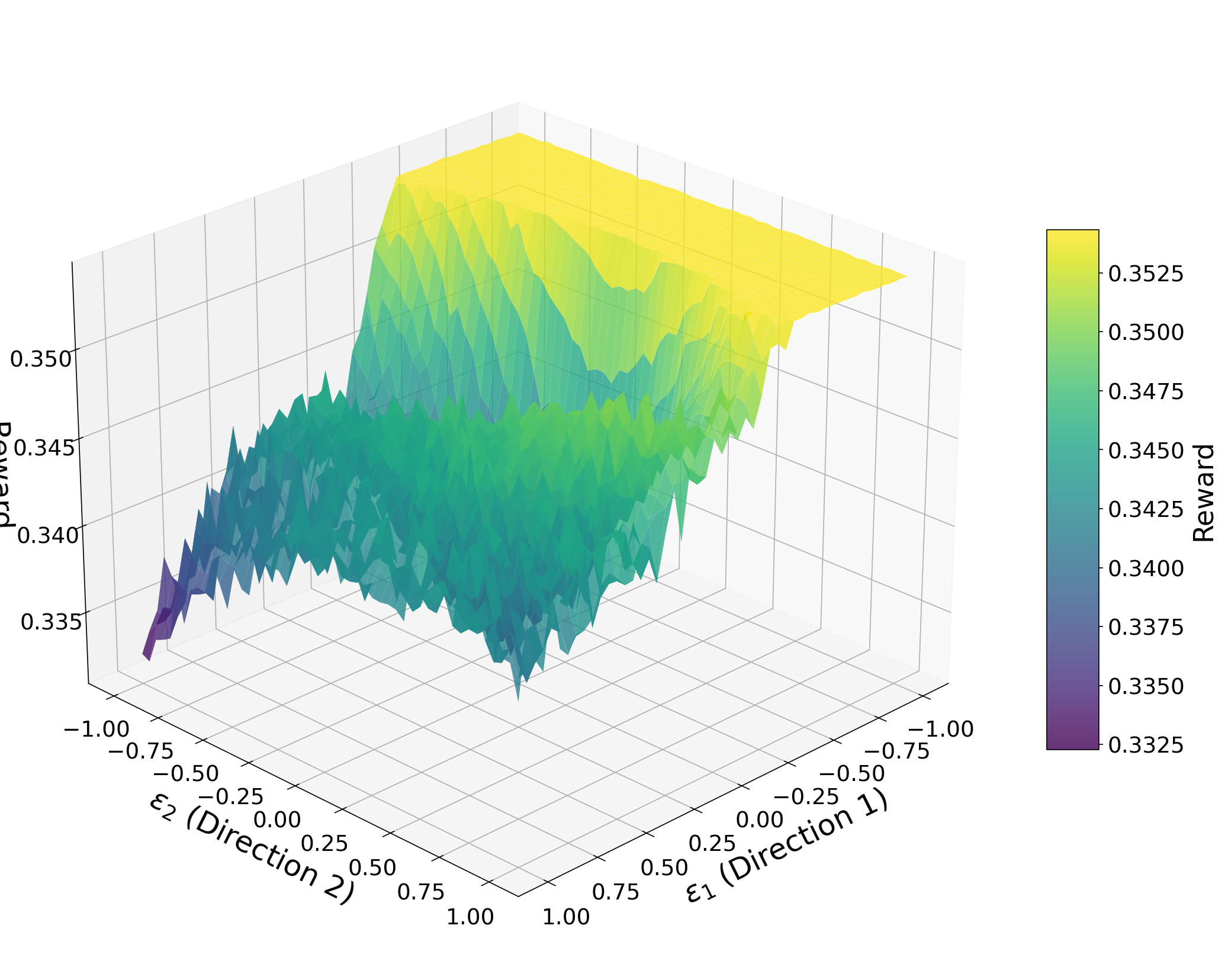}
    \caption{50 adversarial tokens}
    \label{fig:qwen7b_50tok_adv}
  \end{subfigure}

  \vspace{1em}

  \begin{subfigure}[t]{0.49\textwidth}
    \centering
    \includegraphics[width=\linewidth]{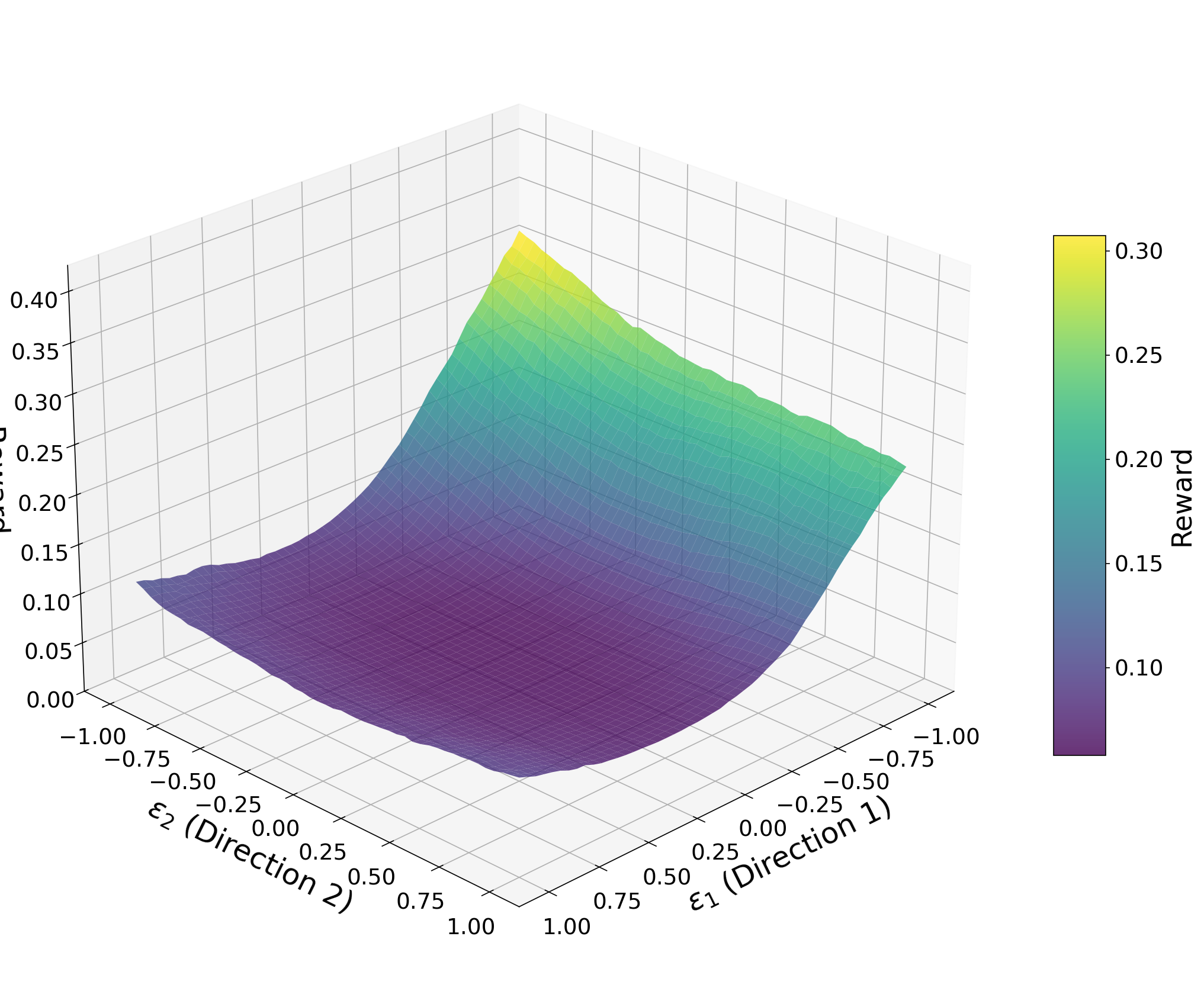}
    \caption{100 random tokens}
    \label{fig:qwen7b_100tok_rand}
  \end{subfigure}\hfill
  \begin{subfigure}[t]{0.49\textwidth}
    \centering
    \includegraphics[width=\linewidth]{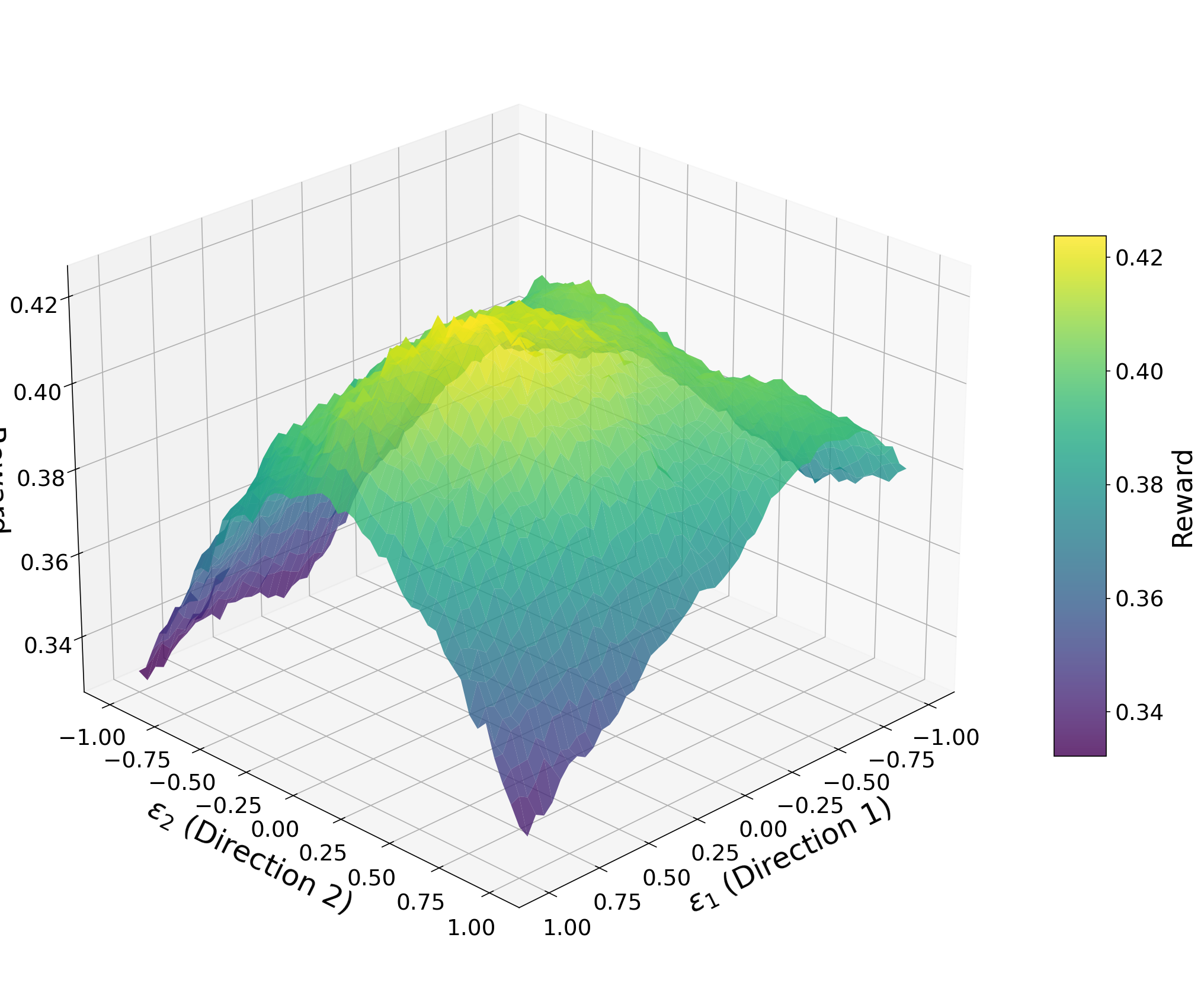}
    \caption{100 adversarial tokens}
    \label{fig:qwen7b_100tok_adv}
  \end{subfigure}

  \caption{Reward landscape visualizations for Qwen-7B: random vs.\ adversarial discrete tokens, averaged across 8 AIME24 trajectories. Note that for Qwen, tokens are inserted between the question and solution rather than appended.}
  \label{fig:qwen7b_landscape}
\end{figure*}

This section provides extended reward landscape visualizations for both PRMs, complementing the analysis in Section~\ref{sec:active_probing}. Figure~\ref{fig:skywork7b_landscape} shows the reward landscapes for Skywork-7B under random and adversarially optimized token sequences appended at the end of trajectories. Figure~\ref{fig:qwen7b_landscape} shows corresponding visualizations for Qwen-7B, where tokens are inserted between the question and solution (middle position) due to Qwen's reward aggregation via minimum. In both cases, adversarially optimized tokens produce more concentrated high-reward regions compared to random baselines, illustrating the exploitability of PRM reward surfaces.

\end{document}